\documentclass{article}

\usepackage{microtype}
\usepackage{graphicx, multicol}
\usepackage{subfigure}
\usepackage{caption}
\usepackage{wrapfig}

\usepackage{booktabs} 
\usepackage{tabularx}
\usepackage{amsmath}
\usepackage{mathabx}

\usepackage{hyperref}
\usepackage{cleveref}

\usepackage[accepted]{icml2021}

\icmltitlerunning{Continuous Coordination As a  Realistic Scenario for Lifelong Learning}
\begin{document}

\twocolumn[
\icmltitle{Continuous Coordination As a  Realistic Scenario for Lifelong Learning}




\icmlsetsymbol{equal}{*}

\begin{icmlauthorlist}
\icmlauthor{Hadi Nekoei}{equal,mila}
\icmlauthor{Akilesh Badrinaaraayanan}{equal,mila,udem}
\icmlauthor{Aaron Courville}{mila,udem,cifar}
\icmlauthor{Sarath Chandar}{mila,poly,cifar}
\end{icmlauthorlist}

\icmlaffiliation{mila}{Mila}
\icmlaffiliation{udem}{Université de Montréal}
\icmlaffiliation{poly}{École Polytechnique de Montréal}
\icmlaffiliation{cifar}{Canada CIFAR AI Chair}
\icmlcorrespondingauthor{Hadi Nekoei}{hadinekoei94@gmail.com}
\icmlcorrespondingauthor{Akilesh Badrinaaraayanan}{akilesh041195@gmail.com}

\icmlkeywords{Lifelong Learning, Multi-agent Reinforcement Learning}

\vskip 0.3in
]



\printAffiliationsAndNotice{\icmlEqualContribution} 

\begin{abstract}
Current deep reinforcement learning (RL) algorithms are still highly task-specific and lack the ability to generalize to new environments. Lifelong learning (LLL), however, aims at solving multiple tasks sequentially by efficiently transferring and using knowledge between tasks. Despite a surge of interest in lifelong RL in recent years, the lack of a realistic testbed makes robust evaluation of LLL algorithms difficult. Multi-agent RL (MARL), on the other hand, can be seen as a natural scenario for lifelong RL due to its inherent non-stationarity, since the agents' policies change over time. In this work, we introduce a multi-agent lifelong learning testbed that supports both zero-shot and few-shot settings. Our setup is based on Hanabi --- a partially-observable, fully cooperative multi-agent game that has been shown to be challenging for zero-shot coordination. Its large strategy space makes it a desirable environment for lifelong RL tasks. We evaluate several recent MARL methods, and benchmark state-of-the-art LLL algorithms in limited memory and computation regimes to shed light on their strengths and weaknesses. This continual learning paradigm also provides us with a pragmatic way of going beyond centralized training which is the most commonly used training protocol in MARL. We empirically show that the agents trained in our setup are able to coordinate well with unseen agents, without any additional assumptions made by previous works. The code and all pre-trained models are available at \href{https://github.com/chandar-lab/Lifelong-Hanabi}{https://github.com/chandar-lab/Lifelong-Hanabi}.
\end{abstract}


\section{Introduction}
Deep reinforcement learning (RL) has shown an immense potential to achieve superhuman performance~\citep{mnih2013playing, silver2018general} on some narrow and well-defined tasks. In contrast, humans can quickly and continually learn new tasks while maintaining the skills to solve previously learned tasks. The ability of an AI system to effectively update new information over time is known as lifelong learning (LLL) or continual learning, and one can postulate this as one of the fundamental ingredients of general AI. Balancing between learning from recent experiences while not forgetting the knowledge acquired from the past is a well-studied problem known as the stability-plasticity dilemma~\cite{carpenter1987massively}. Catastrophic forgetting is a phenomenon in which training the model with new information obstructs previously learned knowledge. This is a common failure case in training neural networks to adapt to new tasks or learning from non-stationary data streams (i.e. non-iid)~\cite{mccloskey1989catastrophic}. Alleviating catastrophic-forgetting is crucial to enable real-world applications where input distributions can shift and where retraining on past data or from scratch is infeasible. While lifelong learning has been identified as an important and challenging problem decades ago~\citep{thrun1998lifelong, ring1998child}, it has recently seen a surge of interest ~\citep{lopez2017gradient, chaudhry2018efficient, chaudhry2019tiny, kirkpatrick2017overcoming, aljundi2018memory} with the success of deep learning.
\par
Several standard benchmarks have been proposed to evaluate novel LLL approaches, mostly for supervised learning settings such as Permuted MNIST~\cite{goodfellow2013empirical}, Split MNIST/CUB/CIFAR~\citep{zenke2017continual, chaudhry2018efficient}. One fundamental issue with using datasets like MNIST as a source of data is the lack of resulting task complexity especially with the large capacity of modern neural networks. Another issue with most current LLL benchmarks is that the relation between tasks cannot be quantified easily. Consequently, most of the evaluation efforts have focused mainly on mitigating catastrophic forgetting, while an ideal LLL system should in addition measure forward and backward transfer. Some recent works have shown limitations of LLL benchmarks~\citep{antoniou2020defining,Roady_2020_Stream51}. For instance, it has been shown that after continual training, the performance of a model trained from scratch using only samples from the episodic memory at test-time, is comparable to specifically designed LLL solutions for most of these benchmarks~\cite{prabhu2020gdumb}. There have been efforts to address this by proposing more challenging benchmarks like CORe50~\cite{lomonaco2017core50}, CRIB~\cite{stojanov2019incremental}, OpenLoris~\cite{shi2020ready}, Stream\-51~\cite{Roady_2020_Stream51}, and IIRC~\cite{abdelsalam2020iirc}. 
\par
RL can be a natural fit for studying LLL as it provides an agent-environment interaction paradigm wherein the agent is exposed to non-stationary streams of data~\citep{kaplanis2018continual, kaplanis2019policy}. However, there is a dearth of well-established benchmarks to study progress in lifelong RL. Most of these benchmarks are hand-engineered customization to the standard RL
environments~\citep{bellemare2013arcade, brockman2016openai} adding synthetic non-stationarity to the environments~\citep{henderson2017benchmark, al2017continuous} or ordering some completely unrelated environments in a sequence~\cite{xu2020taskagnostic} to facilitate the evaluation of LLL performance (eg. a random sequence of Atari used in~\cite{kirkpatrick2017overcoming}). 
Designing overly-tailored experiments for a specific lifelong RL problem can entail unwanted bias in the study~\cite{khetarpal2020towards}.

In this work, we propose a new lifelong RL setup based on Hanabi~\cite{bard2020hanabi} called \textit{Lifelong Hanabi}. Hanabi is a partially-observable, fully cooperative multi-agent game that consists of 2-5 players. In our setup, one agent (\textit{learner}) is trained sequentially with a set of partners (tasks). The \textit{learner} and its partners are sampled from a large pool of pre-trained agents ($\geq$ 100). The pre-trained pool consists of agents trained with different MARL methods such as Independent Q-learning (IQL)~\cite{tan1993multi}, Value Decomposition Networks (VDN)~\cite{sunehag2017value}, Simplified Action Decoder (SAD)~\cite{hu2019simplified}, Other-Play (OP)~\cite{hu2020other} with different architectures and seeds for each method that have shown good performance in Hanabi. \citet{bard2020hanabi} show that agents trained even with the same MARL method but different seeds do not learn to cooperate in the zero-shot scenario, thereby suggesting that these agents converge to different strategies. This large strategy space of Hanabi makes it an ideal scenario for LLL. How far-apart the agents are in the strategy space can be measured through the \textit{cross-play}~(CP) matrix~\cite{bard2020hanabi} that contains the gameplay scores obtained by pairing the agents with one another. CP scores can be used as a proxy for task-similarity to design tasks in \textit{Lifelong Hanabi}. 

\par 

Our contributions are as follows:
\begin{itemize}
    \item We propose a new lifelong reinforcement learning benchmark that has the following desirable properties: (1) It is challenging for state-of-the-art (SOTA) lifelong learning algorithms, (2) It is straightforward to quantify the relation between tasks through the CP matrix, and (3) It is easily extendable to long sequences of diverse tasks without any synthetic modifications. 
    \item We evaluate recent LLL algorithms on this benchmark in limited memory and computation regimes and highlight their strengths and limitations.
    \item We obtain comparable performance on zero-shot coordination in Hanabi even when coordinating with agents trained with MARL methods different from that of the \textit{learner}, without any additional assumptions such as exploiting handcrafted symmetries~\cite{hu2020other} or having access to other agent's greedy action or policy ~\cite{hu2019simplified}.
\end{itemize}


\section{Related work}
In this section, we will provide an overview of existing lifelong RL benchmarks. We will also review recent MARL algorithms since our benchmark is based on a challenging MARL problem.

\subsection{Lifelong Reinforcement Learning Benchmarks}

With regard to lifelong RL benchmarks, ~\citet{henderson2017benchmark} proposed 50 new variations to OpenAI Gym environments through modifying some aspects of either the environments or agents like gravity, morphology of the agent's body, or goal positions. \citet{al2017continuous} introduced RoboSumo — a 3D environment based on MuJoCo physics simulator that allows pairs of agents to compete against each other. The robots differ in anatomy: the number of legs, their positions, and constraints on the thigh and knee joints. \citet{lomonaco2020continual} designed CRLMaze based on VizDoom~\cite{kempka2016vizdoom}, an object-picking LLL task that is composed of 4 scenarios (Light, Texture, Object, All) of incremental difficulty and a total of 12 maps. While these are interesting benchmarks, they still need synthetic modifications to either the environment or the agent in order to introduce non-stationarity. Recently, Coinrun~\cite{cobbe2019quantifying} was proposed that is a procedurally generated environment having different training and testing sets to measure generalization in RL. Jelly Bean World (JBW)~\cite{platanios2020jelly} is a testbed introduced to develop agents with never-ending learning capabilities. It provides support to create non-stationary environments with wide range of tasks including multi-task and multi-modal settings.

\subsection{Multi-agent RL}
In recent years, there has been rapid progress on novel MARL algorithms that are based on centralized training with decentralized execution (Self-Play) as a training paradigm ~\citep{sunehag2017value, foerster2017learning, hong2017deep, foerster2019bayesian, hu2019simplified}. \citet{sunehag2017value} use a Value Decomposition Network (VDN) to learn to decompose the joint state-action value into agent-wise Q-values that depend on local observations of each agent. Bayesian Action Decoder (BAD)~\cite{foerster2019bayesian} and its simplified version (SAD)~\cite{hu2019simplified} propose \textit{public belief} MDP and use an approximate Bayesian update to tackle partially observable tasks. \citet{hong2017deep} aim to learn models of opponents by learning policy features from raw observations of other agents. On the other hand, \citet{omidshafiei2017deep} is one of the few efforts toward decentralized multi-agent learning which shows that optimistic learners can learn sample-efficient MARL policies. There have also been advances towards developing more complex MARL challenges. \citet{stone2010ad} first introduced \textit{ad-hoc} team-work challenge as a multi-agent task where autonomous agents need to collaborate with previously unknown teammates on tasks in which they are all individually capable of contributing as team members as well as characteristics of a good \textit{ad-hoc} player. More recently, in the Hanabi challenge~\cite{bard2020hanabi}, the authors introduce \textit{ad-hoc} scenario as a setting with the objective of making the RL agents adapt to play effectively with unknown partners or even humans. However, SP agents learn brittle policies that fail to cooperate in this scenario~\cite{bard2020hanabi}. Cooperative settings are more reflective of real-world scenarios such as autonomous driving and are crucial for human-AI collaboration~\cite{crandall2018cooperating}.  
The closest work to ours from the perspective of zero-shot coordination is Other-Play \cite{hu2020other}, which exploits the symmetry in the environment by training a self-play agent with shuffled observation space. This simple but elegant idea when combined with adding an auxiliary task, has been shown to be effective in training agents that are able to coordinate with other agents trained with the same MARL methods. However, other-play is still a self-play strategy and requires the knowledge of symmetry of the game upfront. In this work, we aim to go beyond these restricting assumptions and train a lifelong learner that is able to coordinate with unseen agents while not forgetting to play well with previous partners.


\begin{figure*}[th!]
\begin{center}
\includegraphics[width=.98\textwidth]{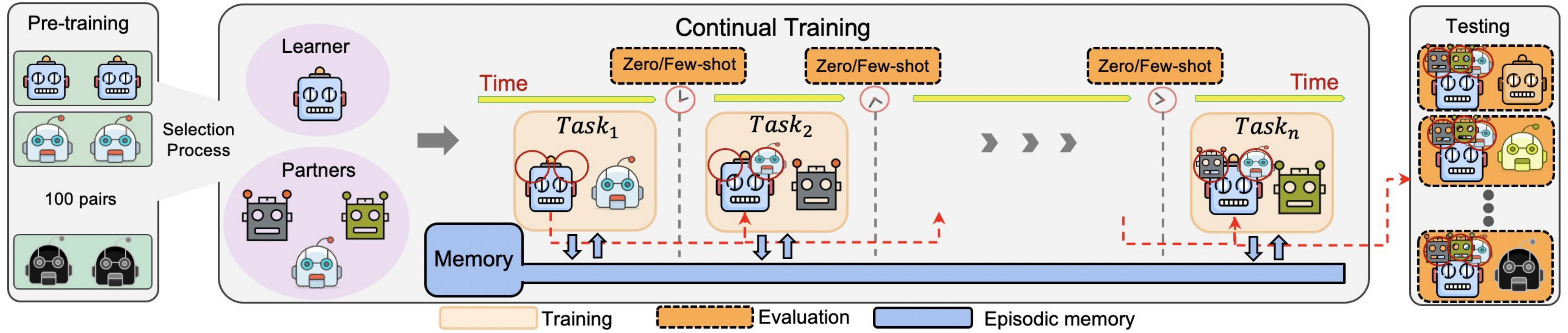}
\caption{Our Lifelong Hanabi setup consists of three phases: \textbf{1- Pre-training (Optional):} In this phase, a pool of agents are trained through SP, \textbf{2- Continual training:} The \textit{learner} is taken from the pool ($\sim p_{train}$) and trained sequentially with some \textit{partners} ($\sim p_{train}$) and periodically evaluated against all the partners, \textbf{3- Testing:} The \textit{learner} is evaluated with a set of random agents excluding its partners ($\sim p_{test}$) to measure generalization.}
\label{fig:setup} 
\end{center}
\vskip -0.2in
\end{figure*}

\section{Multi-Agent RL and Lifelong Learning}

\textbf{MARL for LLL:} Many machine learning algorithms make the assumption that the observations in the dataset are independent and identically distributed (i.i.d). However, in many real-world scenarios, this assumption is violated because the underlying data distribution is non-stationary. Lifelong learning tries to address this problem, where the non-stationarity of data is usually described as a sequence of distinct tasks. On the other hand, MARL is inherently non-stationary due to changing behavior of other agents present in the environment. Therefore, MARL is a realistic scenario for LLL. Another source of non-stationarity in MARL arises when the agent has to interact with different agents over its lifetime, even if the other agents are fixed. For example, in the game of Hanabi, we are interested in designing a single agent that can learn to coordinate well with a sequence of agents it will see over its lifetime. This is a lifelong learning problem.

\textbf{LLL for MARL:}
Standard MARL methods typically focus on the centralized training with decentralized execution setting where agents have access to other agents' policies and observations during training~\cite{zhang2019multi}. Self-Play (SP)~\cite{tesauro1995temporal}, the most common centralized training setting involves training a single agent against itself without any extra supervision. While this strategy works very well in competitive settings like playing the game of Go~\cite{silver2016mastering}, in cooperative settings it can produce agents that establish highly specialized conventions that do not carry over to novel partners they have not been trained with~\cite{bard2020hanabi}. In particular, ~\citet{bard2020hanabi} show that even though RL agents achieve a decent score after training in the SP setting, their performance drops off sharply in the zero-shot coordination scenario, with some agents scoring essentially zero. Therefore, SP agents fail to learn robust strategies that facilitate cooperation with \textit{other} agents. 
Lifelong learning provides a natural framework to transfer knowledge from previous experience to future scenarios. Hence, in this paper, we consider lifelong learning as an alternative to self-play in MARL with the hope that lifelong learning algorithms can learn to coordinate well with unseen agents.

\section{Lifelong Hanabi: A Benchmark for Lifelong Reinforcement Learning}
\par

In this section, we introduce our lifelong reinforcement learning benchmark based on Hanabi. Hanabi~\cite{bard2020hanabi} is a partially-observable, fully cooperative multi-agent game that consists of 2-5 players. Each player can observe other players' hands except his/her own, making it a partially observable game. The objective of the game is to form ordered stacks of cards of respective colors (fireworks). The players can communicate with each other implicitly through actions or explicitly through \textit{hints}, which are limited in number. Thus, Hanabi is a challenging game that requires the agent to possess the \textit{theory of mind}~\citep{premack1978does, rabinowitz2018machine} in order to cooperate effectively. The theory of mind is the ability of an agent to see the world through the lens of other agents. Our objective is to design a training paradigm that can learn zero-shot and few-shot coordination with unseen agents. To this end, we take inspiration from the recent Invariant Risk Minimization (IRM) \cite{arjovsky2019invariant} that improves the Out-of-Distribution (OOD) generalization by training an algorithm on multiple environments. MARL provides such an environment naturally without any need to hand-engineer different features to come up with a set of diverse environments.

As shown in Figure~\ref{fig:setup}, in our Lifelong RL setup, the \textit{learner} ($\sim p_{train}$) is trained sequentially on a set of tasks \textbf{M} = $\{M_{t}\}_{t=1}^T$ sampled from a distribution $p_{train}$ over diverse strategies that perform well in Hanabi. The objective of the \textit{learner} is to learn to coordinate well with its partners during continual training, with the ultimate goal of learning to coordinate well with unseen agents at the end of the training. During the testing phase, in order to measure generalization performance, the \textit{learner} is evaluated with some random agents sampled from $p_{test}$. Although we consider a fully cooperative game in this work, our proposed lifelong RL setup can be easily extended to other multi-agent scenarios (e.g. fully competitive, mixed cooperative-competitive, etc.). Our proposed setup consists of three phases: (1) Pre-training, (2) Continual-training, and (3) Testing. The detailed description of each phase is as follows:

\textbf{Pre-training:}\label{sec:pre-train}
In this phase, agents are trained through SP to play the game of Hanabi. We consider several recent MARL methods for training the agents (IQL/VDN/OP/SAD, and their combinations) across different seeds and architectures leading to a pool of agents having diverse strategies.

\textbf{Continual-training:}
An agent sampled from the pool is chosen as the \textit{learner} and is trained sequentially with a set of agents (partners) for a fixed number of games per partner. The \textit{learner} is also periodically evaluated with all its partners under both zero-shot and few-shot settings. In order to implement memory-based LLL algorithms (ER, A-GEM, etc.), we also include an episodic memory that is used to store some transitions from every task, which can be then be used for replaying in the future tasks. 

\textbf{Testing:} The \textit{learner} is evaluated with $K$ random agents sampled from the pool excluding its partners in order to measure the generalization performance.

\subsection{Evaluation methods} \label{Eval_methods_sec}

We consider two modes of evaluation in our setup : (a) zero-shot and (b) few-shot. In zero-shot setting, the \textit{learner} is evaluated with another agent without providing it a chance to make updates to its own policy through its interaction with the other agent. On the other hand, in the few-shot setting, the \textit{learner} plays a few games with the other agent, thereby adapting its policy through interaction, before being evaluated. We believe agents performing well under both these evaluation settings are crucial to developing AI systems that can adapt well to unknown partners not just in Hanabi but can also facilitate effective collaboration with humans. Few-shot evaluation setting also opens a door to explore recent advances in Meta-RL algorithms to enable fast adaptation.

\subsection{Metrics}
We measure the \textit{learner}'s performance during continual training with some standard metrics from the LLL literature such as average score ($A$), forgetting ($F$), and forward transfer ($FT$)~\citep{lopez2017gradient,chaudhry2018riemannian}. We also define a metric for measuring OOD generalization in our setup called generalization improvement score ($GIS$), inspired by \cite{zhang2018dissection}. To calculate these metrics, we map Hanabi scores at the end of the game from [0, 25] to [0, 1] to have values more consistent with the literature.

\textbf{Average score} ($A \in [0, 1]$):
Let $a_{i,j}$ be the score of the \textit{learner} versus the $j^{th}$ partner, after training it with the $i^{th}$ partner in sequential training. The average score of the \textit{learner} at task $t$ ($A_{t}$) is defined as:
\begin{equation}
    A_t = \frac{1}{t}\sum_{j=1}^{t} a_{t,j}
\end{equation}

\textbf{Forgetting}($F \in [-1, 1]$): Let $f_j^t$ represent the  forgetting on task $j$ after the \textit{learner} is trained on task $t$ and is computed as:
\begin{equation}
    f_j^t = \max_{l \in {1,...,t-1}} a_{l,j} - a_{t,j} \\
\end{equation}
The average forgetting at task $t$ is then defined as:
\begin{equation}
    F_t = \frac{1}{t-1}\sum_{j=1}^{t-1} f_j^t
\end{equation}

\textbf{Average future score} ($FT \in [0, 1]$):
Let $a_{i,j}$ be the score of the \textit{learner} versus the $j^{th}$ partner, after training it with the $i^{th}$ partner in sequential training. The average future score (or forward transfer) of the \textit{learner} at task $t$ ($FT_{t}$) is defined as:
\begin{equation}
    FT_t = \frac{1}{T - t}\sum_{j=t+1}^{T} a_{t,j}
\end{equation}

\textbf{Generalization Improvement Score ($GIS \in [0, 1]$)}: \citet{zhang2018dissection} define in-task generalization as the difference in the performance of an RL algorithm between a set of training and testing trajectories all generated by the same simulator. In this case, the only source of variation is through random seeds. However, in our work, we are more interested in the out-of-task generalization that can be defined as follows. Let $a_{0,k}$ and $a_{N,k}$ be the score of the \textit{learner} versus the $k_{th}$ random agent sampled from the pool (different from its partners) before the start of LLL and at the end of continual training, respectively ($T$ is number of tasks in LLL). The GIS is computed as follows:

\begin{equation}
    GIS = \frac{1}{K} \sum_{k=1}^{K} (a_{T,k} - a_{0,k})
\end{equation}
where $K$ is the total number of unseen agents.

\section{Experiments}

\begin{figure}[t]

\centering
\includegraphics[trim = 0 0 0 0, clip, width=0.47\textwidth]{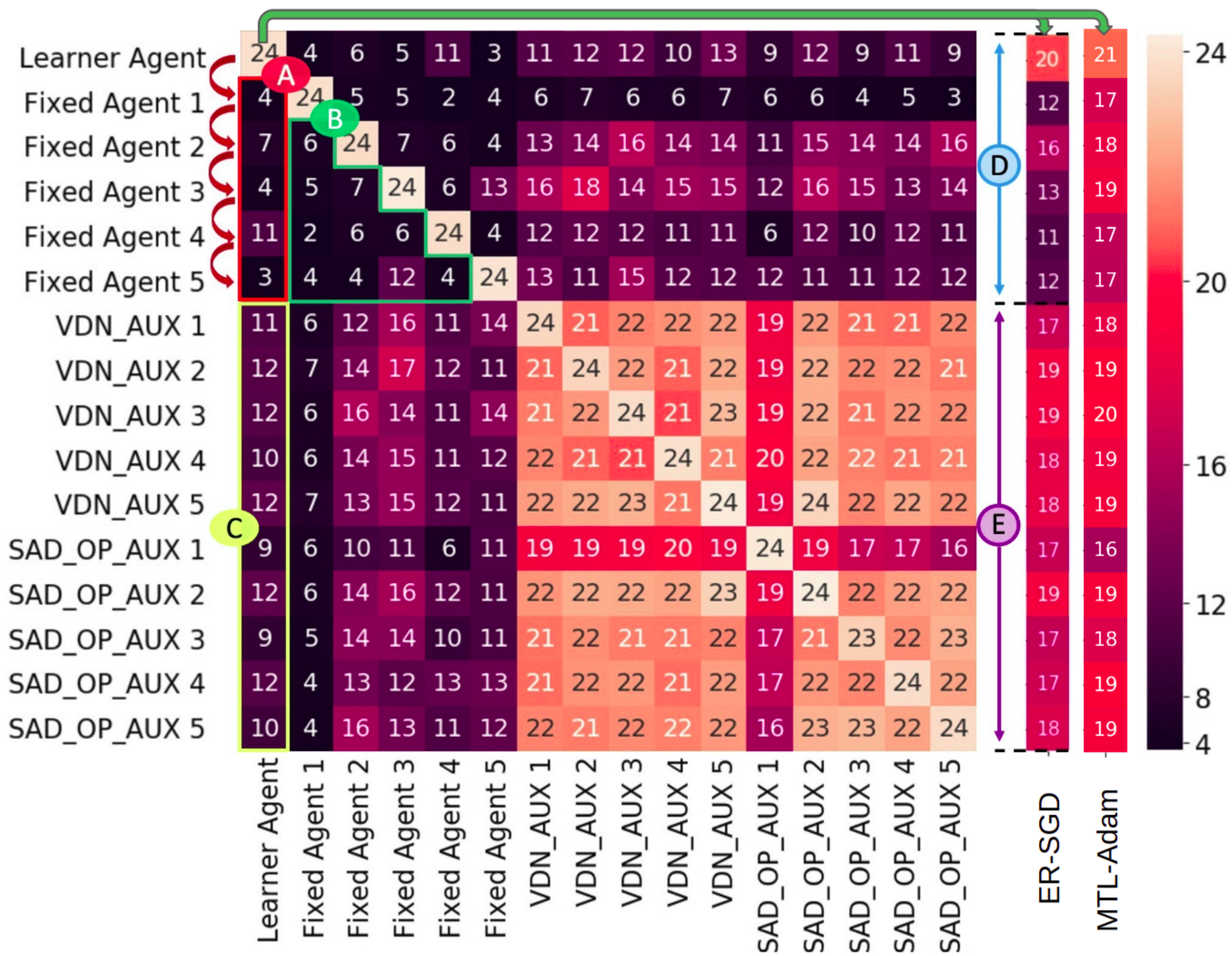}

\caption{CP scores before (left) and after continual training (right) -- $(i, j)_{th}$ element is the average score of agent $i$ paired with $j$. [A-C] is before continual training -- (A) Initial scores of the \textit{learner} with its partners, (B) Cross-play scores amongst the partners, low scores indicate they are far apart in the strategy space, (C) Initial generalization scores with some unseen agents. The \textit{learner} is then trained continually with its partners following the order indicated by the arrows. [D-E] is after continual training -- (D) Scores with the original learner and its partners, (E) Generalization scores with the same unseen agents.}
\label{fig:motivation}
\end{figure}

Figure~\ref{fig:motivation} showcases how continual training can lead to improved scores and better zero-shot coordination. An IQL agent (pre-trained with self-play) is trained sequentially with 5 partners (ordering denoted by arrows in Figure~\ref{fig:motivation}), and then evaluated with both its partners seen during continual training as well as some unseen agents (SAD+AUX+OP, VDN+AUX). Sections (C) and (E) in Figure~\ref{fig:motivation} show the performance of the IQL agent on unseen agents before and after continual training respectively, clearly indicating OOD generalization. Likewise, sections (A) and (D) show the performance before and after training respectively with its partners used in continual training.    

As described in Section~\ref{sec:pre-train}, we first pre-train a set of agents to play the game of Hanabi through SP. Our RL agents are based on the R2D2 architecture~\cite{kapturowski2018recurrent} that are RNN-based DQN agents. The diversity in strategies learned by these SP agents are controlled by varying the seed, the MARL methods used for training (either IQL/VDN/OP/SAD and their combinations), type of recurrence (LSTM/GRU), number and dimension of recurrent layers, number and dimension of feed-forward layers before recurrence. The exact architecture details are described in the Appendix~\ref{app:agents_list}. A pool of more than 100 SP agents is created this way. A subset of 100 agents with 10 agents from each of these 10 MARL methods are used to generate the \textit{cross-play} (CP) matrix as shown in Appendix-\ref{app:pool}. The entries in this matrix (diagonal entries indicate self-play scores) are obtained through the gameplay of agents with each other for 5k games, and then averaging the scores.

We propose two levels of tasks based on these scores that have different difficulty levels: \textit{easy} and \textit{hard}. In both the settings, one of the agents is used as \textit{learner} and the rest as its \textit{partners} that are fixed during continual training i.e. they represent different tasks since these agents have different strategies. Both the \textit{learner} and its \textit{partners} are initialized with weights of the pre-trained agents as we found pre-training to be crucial to learn some basic knowledge of Hanabi. The names of these tasks are self-explanatory in the sense the \textit{learner} in the \textit{hard} version has to start from a much lower \textit{cross-play} score and learn to achieve a good final score (out of 25). During continual training, the \textit{learner} is trained sequentially with every \textit{partner} for a fixed number of epochs and evaluated periodically with all its \textit{partners} under both zero-shot and few-shot settings. In the zero-shot setting, the \textit{learner} is directly evaluated with all its \textit{partners}, while in the few-shot setting, the \textit{learner} is fine-tuned with its \textit{partner} for a few gradient steps before evaluating against the same partner. In both these settings, scores are reported based on average over 5k gameplay.

\begin{figure*}[th!]
\centering
\includegraphics[trim=200 0 200 0, width=.99\linewidth]{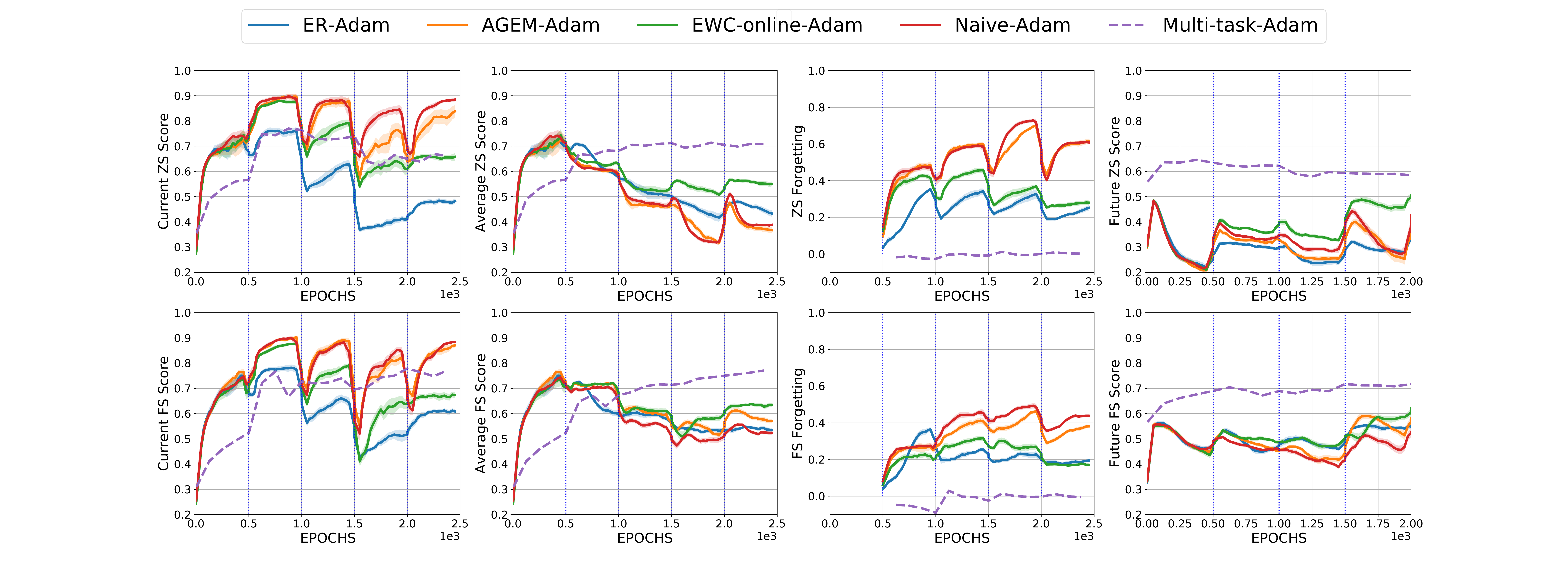}
\caption{Zero-shot (top row) and Few-shot (bottom row) performance of different LLL algorithms with Adam optimizer on \textit{hard} task. From left to right: current score~($\uparrow$), average score~($\uparrow$), forgetting~($\downarrow$), and average future score~($\uparrow$) respectively. ($\uparrow$ = higher better, $\downarrow$ = lower better).}
\label{fig:LLLbenchmark_adam_hard} 
\vskip -0.2in
\end{figure*}

Although the \textit{hard} and \textit{easy} settings consist of five tasks, our setup as such can be effortlessly extended to any number of tasks by selecting a different number of partners from the pre-trained pool and the pre-trained pool can itself be expanded by training more agents through SP. For instance, we report results on 10 tasks in section~\ref{sec:LifelongRL} as well as in Appendix~\ref{app:all_LLL_benchmarks}. Note that to choose the \textit{partners}, we excluded all the agents using either SAD or AUX from the pre-trained pool as we wanted to compare the continually trained learner with them in terms of zero-shot coordination. We also wanted to select partners that have low CP scores so that the tasks are diverse.

R2D2 agents keep recent game transitions in a fixed-size prioritized replay buffer~\cite{schaul2015prioritized}. At the end of every task, the replay-buffer is sliced and stored in an episodic memory, which is then used for replay in different memory-based LLL algorithms that we consider in our benchmark. The \textit{learner} can also start with random parameters (i.e. without pre-training), albeit, this setting is very hard for the game of Hanabi.

We aim to answer the following questions through our experiments : (1) How well standard LLL algorithms perform in our setup (section~\ref{sec:lll-benchmarking}), (2) How well do these LLL algorithms fare under constrained memory and compute settings (section~\ref{sec:constrained_exp}), (3) How lifelong RL methods perform in our setup (section~\ref{sec:LifelongRL}), (4) How well the agents trained in our setup do in zero-shot and few-shot coordination scenarios in Hanabi (section~\ref{sec:marl_comparision}), when compared to other recent methods such as OP~\cite{hu2020other}.

\subsection{Lifelong learning benchmarking:}
\label{sec:lll-benchmarking}
We implement some standard LLL algorithms that are both replay-based and regularization-based. 

\textbf{Naive:} This is the simplest algorithm in which the \textit{learner} is trained sequentially on subsequent tasks, starting with the learned parameters at the end of the previous task, without any episodic memory or regularization.

\textbf{Experience Replay (ER):}
We follow the procedure described in ~\cite{chaudhry2019tiny} for implementing ER. We sample a mini-batch $B_{k}$ from the current task (in which the \textit{learner} plays with partner $k$), and a mini-batch $B_{m}$ that consists of an equal number of samples from all previous tasks collectively. These mini-batches are stacked and a single gradient step is used to update the \textit{learner}. Our implementation closely resembles the ring buffer strategy described in~\cite{chaudhry2019tiny}, as there's equal representation from all previous tasks when sampling $B_{m}$, although the samples within each task itself are prioritized.  

\textbf{Averaged Gradient Episodic Memory (A-GEM):}
Minibatches $B_{k}$ and $B_{m}$ are sampled as described in~\cite{chaudhry2018efficient} and similar to ER. The gradients corresponding to these mini-batches are first computed denoted by $g$ and $g_{ref}$ respectively. If $g^{T}g_{ref} \geq 0 $, then the gradient of the current task $g$ is directly used to update the \textit{learner}'s parameters, whereas if $g^{T}g_{ref} < 0 $, $g$ is first projected such that $g^{T}g_{ref} = 0$ before updating the \textit{learner}. This projection ensures that the average loss over the previous tasks does not increase. 

\textbf{Elastic Weight Consolidation (EWC):}
EWC is a regularization-based technique proposed to alleviate catastrophic forgetting by selectively reducing the plasticity of weights drawing inspiration from Bayesian methods~\cite{kirkpatrick2017overcoming}. EWC uses Fisher information matrix as a surrogate for the importance of learned weights and uses that for gradient updates. Offline EWC uses one Fisher matrix per task, therefore the number of regularization terms increases linearly with the number of tasks whereas Online EWC~\cite{schwarz2018progress} uses only one Fisher matrix that is computed based on all the previous tasks. We consider both these variants in our benchmark.

\textbf{Stable naive/ER/A-GEM/EWC:}
~\citet{mirzadeh2020understanding} show that catastrophic forgetting can be mitigated through careful design of training regimes such as learning rate decay, batch size, dropout, and optimizer that can widen the tasks' local minima. The resulting model with these optimal choices is referred to as ``stable''. In particular, we consider if using larger batches, exponential learning rate decay, dropout (either in feed-forward or recurrent layers in R2D2), and SGD optimizer help improve continual training, thereby leading to better final performance as well as generalization to unseen agents. 

\begin{table}[!h]
\begin{center}
\begin{small}
\caption {Benchmarking LLL methods on Hanabi. Average accuracy and forgetting of LLL algorithms on \textit{hard} task averaged over 5 runs with 5000 games. ($\uparrow$ = higher better, $\downarrow$ = lower better)} 
\label{tab:LLLbenchmark}
\vskip 0.15in
\begin{tabularx}{1.\linewidth}{lcclcc}
\toprule
 Method &  \multicolumn{2}{c}{Zero-shot} & & \multicolumn{2}{c}{Few-shot} \\ \cline{2-3} \cline{5-6}
    &  $A_T \uparrow$  & $F_T \downarrow$       
	&  & $A_T \uparrow$  & $F_T \downarrow$  \\ \cline{2-3} \cline{5-6}
	\addlinespace[3pt]
 Naive Adam &  $0.39$ & $0.60$ & & $0.52$ & $0.44$ \\
EWC off. Adam  & $0.45$ & $0.45$ & & $0.60$ & $0.27$ \\
EWC on. Adam  & $\mathbf{0.55}$ & $0.28$ & & $\mathbf{0.63}$ & $0.17$ \\
ER Adam  & $0.44$ & $0.26$ & & $0.53$ & $0.20$ \\
AGEM Adam  & $0.38$  & $0.62$ & & $0.57$ & $0.38$\\
\midrule
Naive SGD  & $0.52$ & $0.15$ & & $0.52$ & $0.14$ \\ 
EWC off. SGD  & $0.49$  & $0.12$ & & $0.50$ & $0.11$ \\ 
EWC on. SGD  & $0.47$  & $0.12$ & & $0.48$ & $0.11$ \\

ER SGD  & $0.50$  & $\mathbf{0.05}$ & & $0.50$ & $\mathbf{0.04}$ \\
AGEM SGD  & $0.50$  & $0.13$ & & $0.51$ & $0.12$\\
\midrule
Multi-Task Adam & $0.70$ & $0$ & & $0.77$ & $0$ \\
 Multi-Task SGD & $0.50$ & $0$ & & $0.51$ & $0$\\
\bottomrule
\end{tabularx}
\end{small}
\end{center}
\end{table}

\textbf{Multi-Task Learning (MTL):}
In this setting, there's a common replay buffer that contains the experiences of the \textit{learner} interacting with all its partners. Mini-batches sampled from this common replay buffer are used for training the \textit{learner}. This serves as an upper-bound on the achievable performance in our benchmark. 

\par
\begin{figure}[h!]
\centering
\begin{subfigure}[Adam vs SGD]{
    \centering
    \includegraphics[trim=50 50 50 50, width=0.465\linewidth]{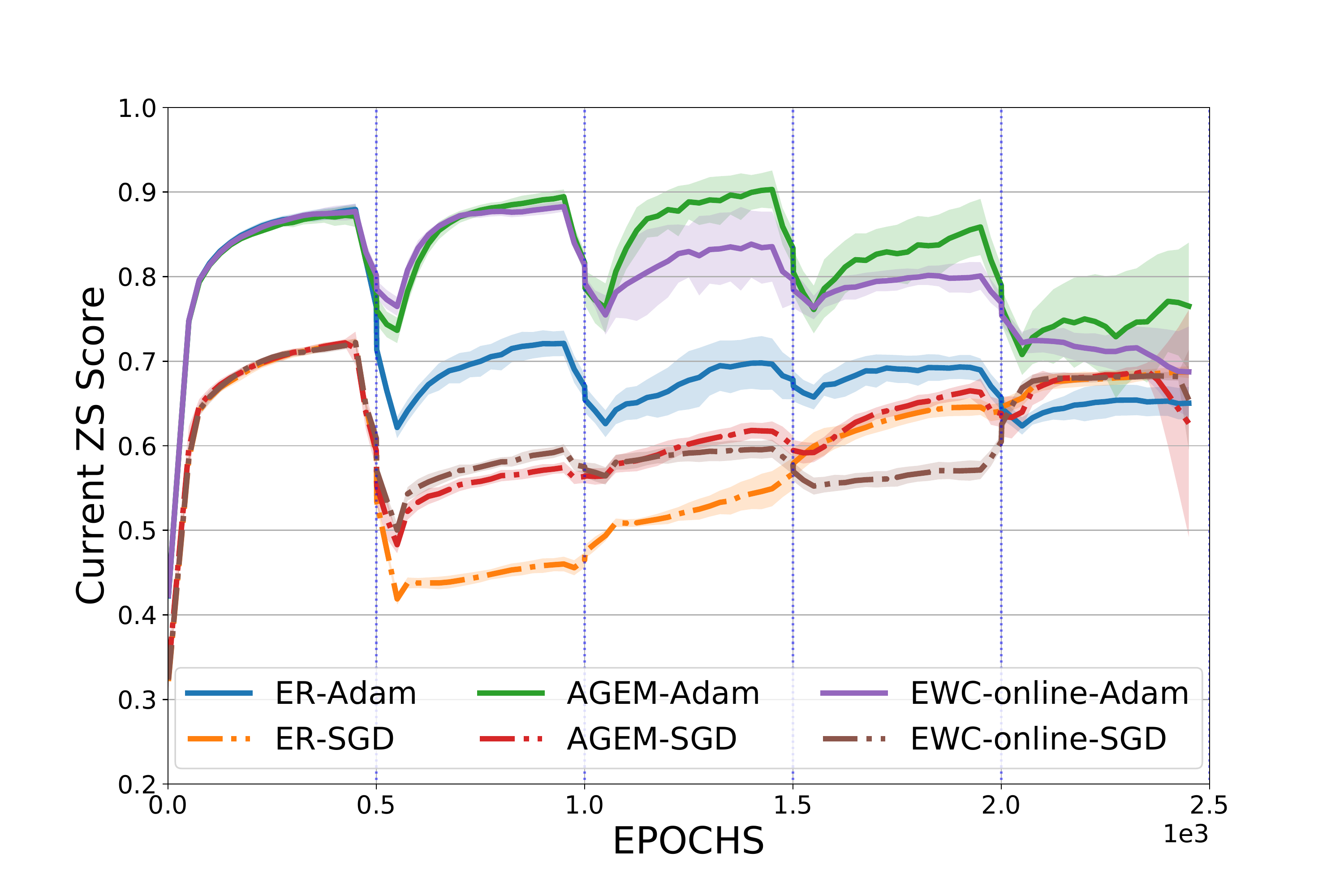}
    }
    \label{fig:Adam_SGD}
\end{subfigure}%
\begin{subfigure}[GIS]{
    \centering
    \includegraphics[trim=50 50 50 75, width=0.46\linewidth]{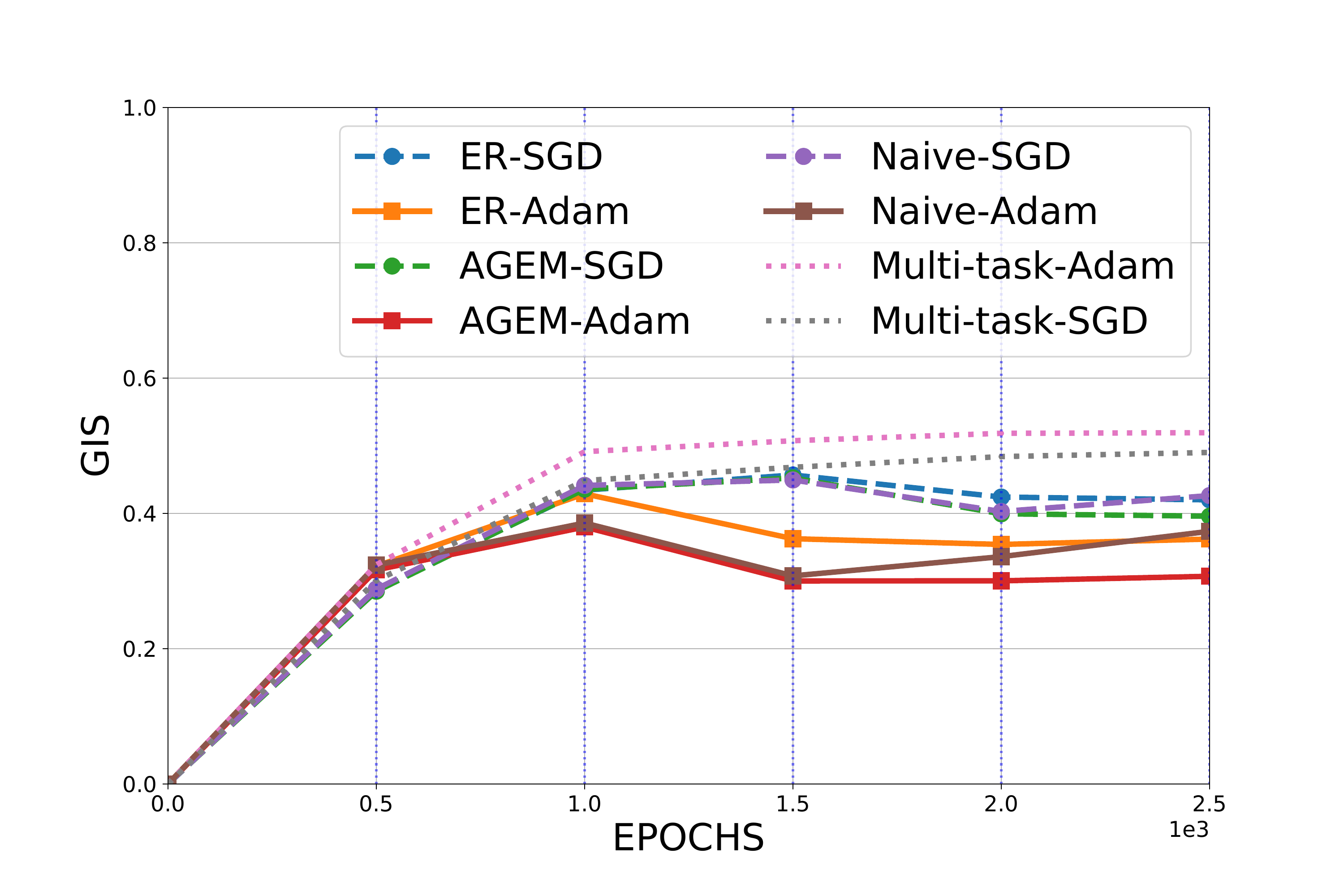}
    }
    \label{fig:GS}
\end{subfigure}%
\begin{subfigure}[Episodic memory size]{
    \centering
    \includegraphics[trim=50 0 50 75, width=0.465\linewidth]{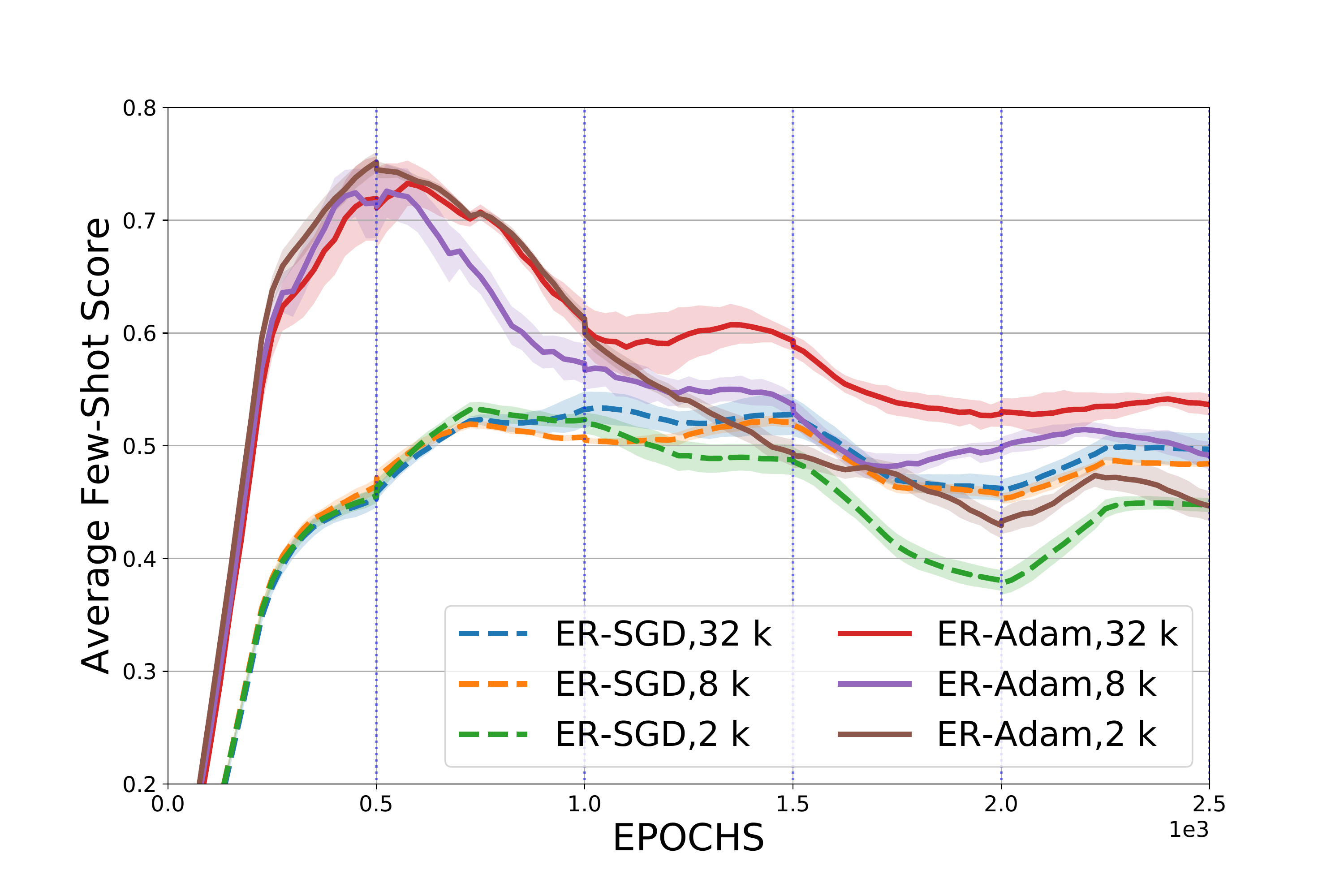}
    }
    \label{fig:Episodic_memory_size}
\end{subfigure}%
\begin{subfigure}[Few-shot gradient steps]{
    \centering
    \includegraphics[trim=50 0 50 100, width=0.46\linewidth]{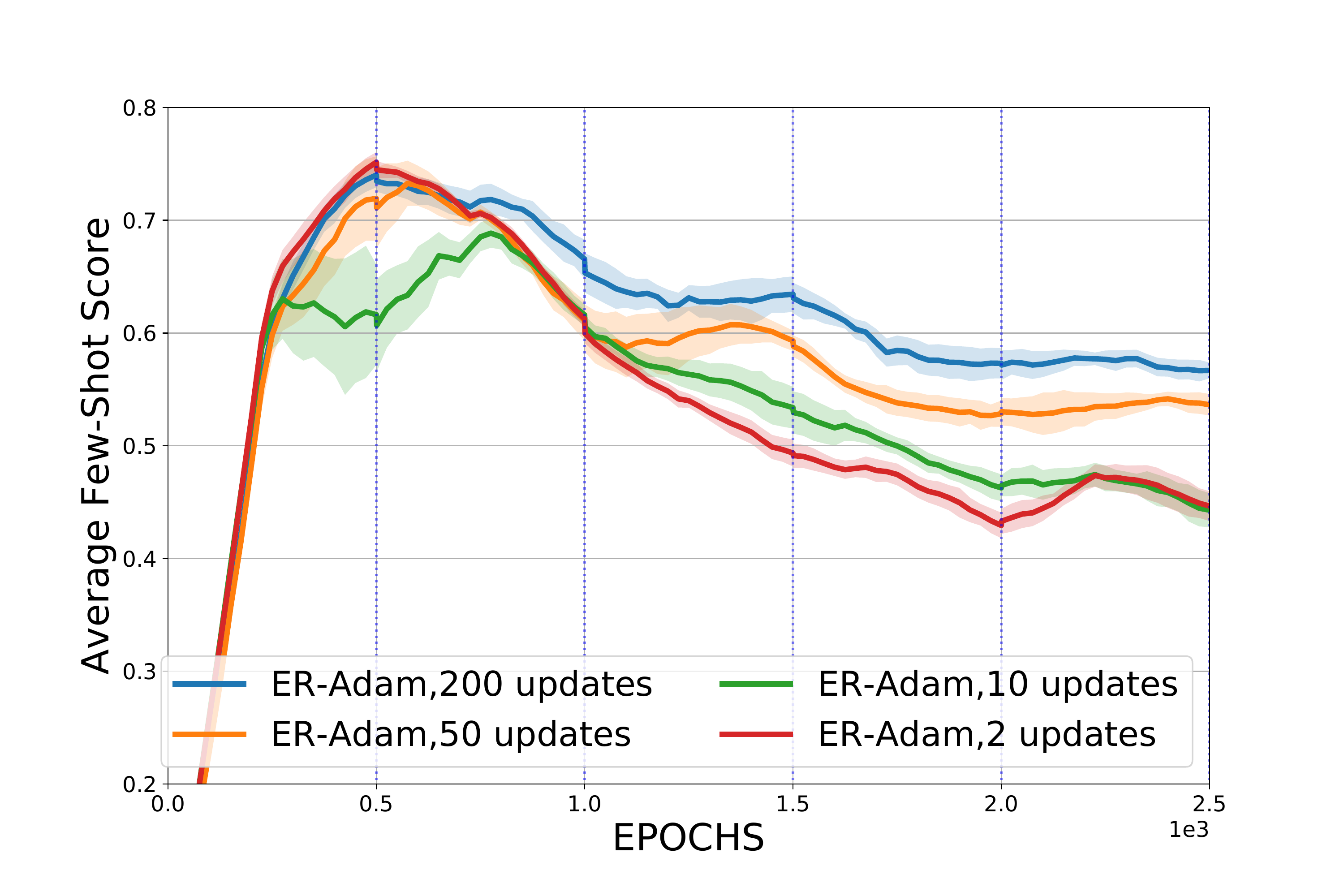}
    }
    \label{fig:few_shot_gradient_steps}
\end{subfigure}%
\caption{More experiments: Generalization score with Inter-CP agents at the end of every task during continual training. LLL algorithms using SGD as an optimizer have better generalization performance compared to Adam. ER SGD has the highest GIS.}
\label{fig:ablation} 
\vskip -0.1in
\end{figure}

 We can observe from Figure~\ref{fig:LLLbenchmark_adam_hard} and Table~\ref{tab:LLLbenchmark} that online EWC with Adam has the best average score in both the zero-shot and few-shot setting among the LLL algorithms, while the forgetting is least for ER with SGD. Table~\ref{tab:LLLbenchmark} also shows the effect of the optimizer on different LLL algorithms. LLL algorithms with SGD tend to have comparatively less forgetting and better zero-shot performance on average. We can infer from  Figure~\ref{fig:ablation} that using Adam helps in fast adaptation to current task albeit at the expense of greater forgetting. This effect can also be seen in higher average few shot scores (Table~\ref{tab:LLLbenchmark}). As one might expect, MTL with Adam achieves the highest average score in both zero-shot and few-shot. 

From the last column of Figure~\ref{fig:LLLbenchmark_adam_hard}, we can observe that when the \textit{learner} start playing with a new partner, there is an increase in average future score suggesting it has learned some useful skills from previous tasks that is transferable to the other partners. However, with more training, the \textit{learner} possibly overfits to its current partner, leading to a drop in average future scores. 

To explore the effect of different training regimes on continual training, we study the effect of larger batches (128 vs 32), learning rate decay with high initial rates (0.2/0.02), and dropout. 
Our experiments suggest that \textit{Lifelong Hanabi} does not benefit greatly from the use of large batches as the gain in scores is negligible. This observation aligns with the ``stable'' networks~\cite{mirzadeh2020understanding} that suggests using small batches. In the case of EWC, we find that there is a stark difference in performance with different $\lambda$ values (the weight assigned to the Fischer term). Our experiments indicate larger $\lambda$ is beneficial. Appendix-~\ref{app:all_LLL_benchmarks} contains the training curves of LLL algorithms for other settings --- \textit{hard} task with SGD as well as \textit{easy} task with both Adam and SGD.

\subsection{Lifelong learning under constrained memory and compute:}\label{sec:constrained_exp}
\subsubsection{Episodic memory size}
In order to understand the effect of episodic memory on the performance of memory-based LLL algorithms, we vary the episodic memory size (\{2k, 8k, 32k\} $\times$ number of tasks) in the case of ER with both SGD and Adam as shown in Figure~\ref{fig:ablation}. In both these cases, a larger episodic memory size results in a better final performance.

\subsubsection{Gradient updates for few-shot evaluation}
To better understand the ability of the \textit{learner} trained with different LLL algorithms to adapt quickly to all its partners, we vary the number of gradient steps used to update the \textit{learner} in the few-shot evaluation scenario. As it can be seen from Figure~\ref{fig:ablation}, there is a considerable difference between 10 and 50 gradient updates on the final performance, whereas the benefit reaped beyond 50 updates is minimal.

\subsection{Lifelong RL methods:} \label{sec:LifelongRL}

\begin{figure}[t!]
\centering
\includegraphics[trim=150 0 150 0, width=.99\linewidth]{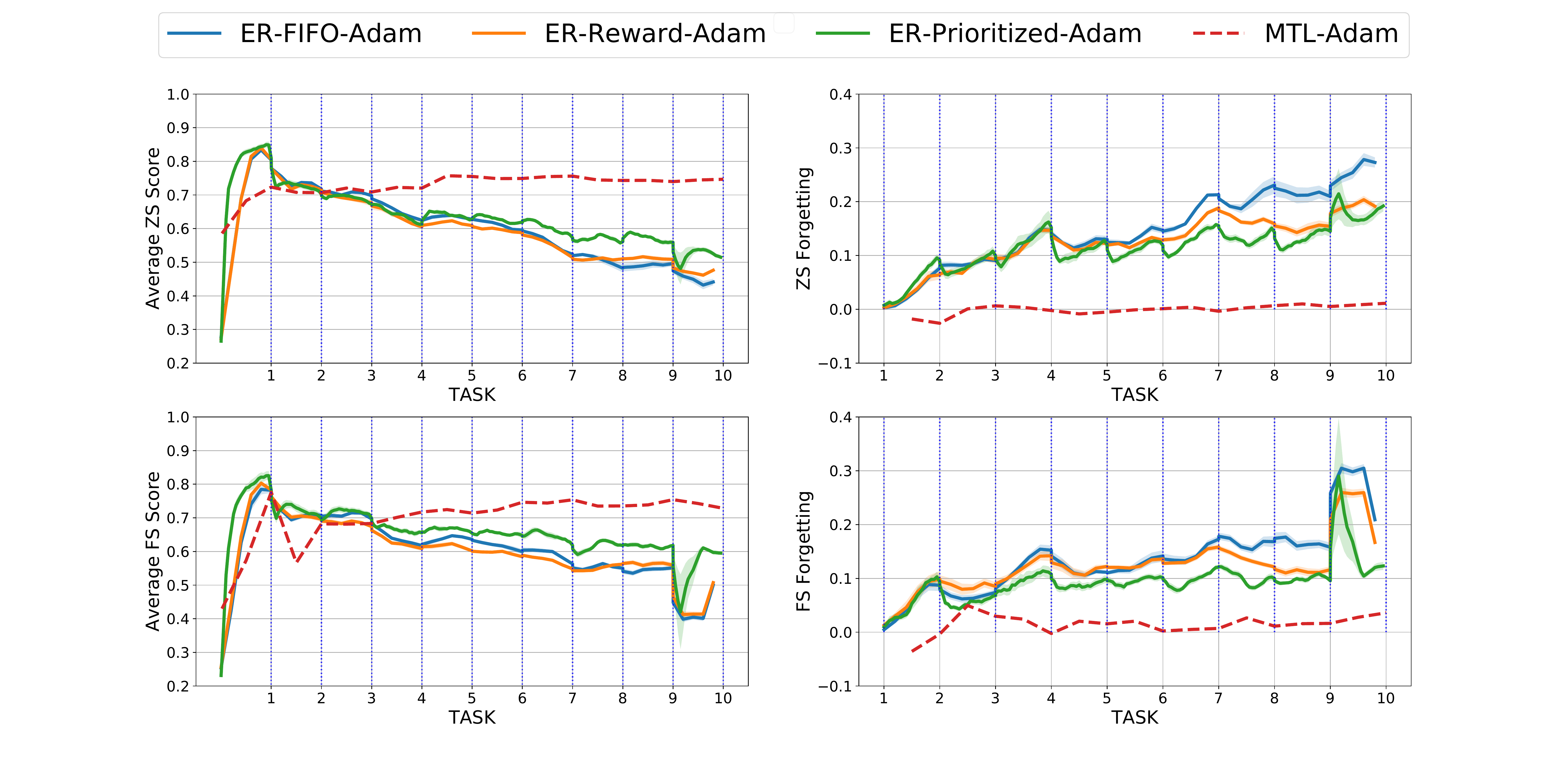}
\caption{Zero-shot (top row) and Few-shot (bottom row) performance of ER methods with different types of episodic memory designed for lifelong RL \cite{isele2018selective} with Adam optimizer on 10 tasks.}
\label{fig:LifelongRL_beanchmark} 
\vskip -0.35in
\end{figure}

\citep{isele2018selective} propose some strategies for storing experiences in the replay-buffer that has been shown to reduce catastrophic forgetting in RL. All our methods use prioritized replay buffer that resembles the \textit{surprise} strategy~\citep{isele2018selective}. In addition, we also compare this with \textit{FIFO} and \textit{Reward} strategies. For \textit{FIFO}, we set the prioritization exponent $\alpha$ to 0 \cite{schaul2015prioritized}, which is equivalent to uniformly sampling. In case of \textit{Reward}, we do prioritized sampling that favors experiences based on the absolute value of reward instead of TD-error as done in our default case. As can be seen in Figure~\ref{fig:LifelongRL_beanchmark}, ER with prioritized sampling performs best compared to \textit{Reward} and \textit{FIFO} strategies in terms of both average score and average forgetting. Implementing other sampling strategies such as \textit{Global Distribution Matching} and \textit{Coverage Maximization} are left for future work.

\begin{table*}[!t]
\caption{Comparison with other MARL algorithms on self-play (SP), cross-play evaluation scores within method (\textit{Intra-CP}), and across different methods (\textit{Inter-CP}). C: centralized training, GA: agents share their greedy action along with their standard action, L: true labels of cards needed, SYM: symmetries of the game needed upfront, P: require access to some pre-trained agents in sequence, UP: Having access to all the fixed pre-trained agents at the same time. ($\uparrow$ / $\downarrow$ = Difference in score after continual training, \textcolor{red}{red}: pre-trained with MARL method, \textcolor{blue}{blue}: trained continually with LLL method)}
\label{sota-marl-table}
\vskip 0.15in
\begin{center}
\begin{small}
\begin{tabularx}{1.\linewidth}{lcccccccc}
\toprule
Training Method && SP & & Intra-CP &&  Inter-CP & Limitations \\
\midrule
\textcolor{red}{SAD}     && $23.85 \pm 0.03$ & & $7.70 \pm 0.69$ && $14.60 \pm 0.24$ & C + GA \\
\textcolor{red}{SAD} + \textcolor{red}{AUX}  && $23.57 \pm 0.03$ & & $20.97 \pm 0.80$ & & $18.51 \pm 0.23$ &   C + GA + L\\ \textcolor{red}{SAD} + \textcolor{red}{OP}  && $24.14 \pm 0.03$ & & $10.10 \pm 0.87$ & & $16.09 \pm 0.25$ & C + Sym + GA \\
\textcolor{red}{SAD} + \textcolor{red}{AUX} + \textcolor{red}{OP}  && $23.40 \pm 0.04$ & & $21.23 \pm 0.25$ &&  17.77 $\pm$ 0.23 &  C + Sym + L + GA       \\
\midrule
\textcolor{red}{IQL} + \textcolor{blue}{ER}  && $20.91 \pm 0.05$ ($\downarrow 2.98$) && $15.73 \pm 0.39$ ($\mathbf{\uparrow 7.06}$)  && $16.32 \pm 0.21$ ($\mathbf{\uparrow 8.09}$)  & P \\
\textcolor{red}{IQL} + \textcolor{red}{AUX} + \textcolor{blue}{ER}  &&  $22.34 \pm 0.06$ ($\downarrow 1.46$)  && $20.90 \pm 0.06$ ($\downarrow 0.15$)  && $\mathbf{19.17 \pm 0.22}$ ($\mathbf{\uparrow 1.33}$)  & L + P \\
\textcolor{red}{IQL} + \textcolor{blue}{Multi-task}   && $20.93 \pm 0.09$ ($\downarrow 2.96$)  && $16.05 \pm 0.30$ ($\mathbf{\uparrow 7.38}$)  && $\mathbf{17.88 \pm 0.17}$ ($\mathbf{\uparrow 9.65}$)  &  UP  \\
\bottomrule
\end{tabularx}
\end{small}
\end{center}
\vskip -0.1in
\end{table*}

\subsection{Zero-shot coordination:} \label{sec:marl_comparision}

We compare our best-performing LLL algorithms with recent MARL methods that have shown good performance on Hanabi (Table~\ref{sota-marl-table}). In addition to reporting self-play evaluation scores, we evaluate each training method with two sets of unseen partners under zero-shot coordination scenario: (1) \textit{Intra-CP} - a set consisting of agents that are trained with the same MARL method as the training method. For example, the SAD+OP agent is evaluated with other SAD+OP agents only, but with different architectures and seeds. Similarly, in order to evaluate agents trained in our setup, we evaluate them with other agents that are trained with the same MARL method as the \textit{learner}, (2) \textit{Inter-CP} - a set containing 20 agents across all the MARL methods we consider. 

As we can observe from Table~\ref{sota-marl-table}, although recent MARL methods can achieve good scores in SP and \textit{Intra-CP} evaluations, they fail to achieve high scores in \textit{Inter-CP} highlighting their inability to coordinate effectively with other MARL methods in the zero-shot scenario. We can observe that agents trained in our setup have significant improvement in both \textit{Inter-CP} and \textit{Intra-CP} compared to the agent at the start of continual learning, however, their SP scores are lower than at the start. The difference in scores due to continual training is indicated in brackets. It is also worth mentioning that IQL+AUX+ER achieve a better \textit{Inter-CP} score than even other MARL methods, although this comes at a cost of slight reduction to \textit{Intra-CP} score.

All the training methods in Table~\ref{sota-marl-table} have some limitations that we highlight now. During training, SAD allows agents to have access to the greedy action of their team mates in addition to the actual exploratory action chosen ($GA$). AUX refers to having an auxiliary task that predicts the \textit{learner}'s own hand and hence requires ground truth labels for this ($L$). OP requires symmetries of the game to be known beforehand ($SYM$). IQL+ER and IQL+AUX+ER require pre-trained agents in sequence for LLL ($P$), while IQL + Multi-Task requires access to all pre-trained agents simultaneously ($UP$). Figure~\ref{fig:ablation} shows the progression in generalization performance (\textit{Inter-CP}) after every task during continual training for several LLL algorithms. MTL (with Adam) and ER (with SGD) have the best scores with the \textit{Inter-CP} agents at the end of continual training. However, MTL needs to interact with all the partners' at the same time which is not always a realistic assumption.  

\section{Conclusions and future work}
In this work, we proposed \textit{Lifelong Hanabi} as a new challenging benchmark for lifelong RL. The non-stationarity in our benchmark was introduced through agents having different strategies instead of synthetic modifications to the environment or agent, while cross-play score served as an easy metric to quantify the similarity between tasks. We analyzed the performance of some well-known LLL algorithms on this benchmark. We also showed that an IQL agent continually trained in our setup can zero-shot coordinate effectively with unseen agents. We hope that the lifelong RL community adopt this as a standard benchmark for evaluating algorithmic advances due to its ease of use.

\textit{Lifelong Hanabi} aims to facilitate development of novel algorithms for lifelong learning specific to RL (i.e. lifelong RL). This framework also serves as a step towards thinking beyond centralized training in MARL. Some interesting future directions are to understand the kind of policies learned by the agents trained in our setup through policy visualization to see what kind of conventions (if any) emerges. It is also valuable to evaluate our trained agents with humans, as developing artificial agents capable of coordinating effectively with humans is an important long-term goal of modern AI. Moreover, exploiting recent advances in Meta-RL such as~\citep{zintgraf2019fast} for faster adaptation in the \textit{few-shot} evaluation setting, instead of naively fine-tuning can lead to agents that adapt well in the \textit{ad-hoc} scenario. We believe studying the effect of the order of \textit{partners} that the \textit{learner} encounters and its effect on final performance is an interesting next step. Currently, the  non-stationarity across different strategies is what we only exploit to design LLL tasks. This already resulted in an interesting trade-off for the learner between adapting to new partners and not forgetting to coordinate well with previous partners. Our preliminary experiments suggest that extending our framework to learning partners is extremely difficult for current methods, however, this could be an exciting future research direction.

\textbf{Acknowledgment}
We would like to thank Gabriele Prato, Mahta Ramezanian, Khimya Khetarpal, Matthew Riemer, Ankit Vani, Moksh Jain, and Salem Lahlou for reviewing the paper and their valuable feedback. We are also grateful to Hengyuan Hu for patiently answering queries regarding the Hanabi SAD code repository, Darshan Patil and Rodrigo Chavez Zavaleta for reviewing our code and Olexa Bilaniuk for helping us with engineering issues internally at Mila. We would like to acknowledge Compute Canada and Calcul Quebec for providing computing resources used in this work. SC is supported by a Canada CIFAR AI Chair and an NSERC Discovery Grant.

\bibliography{ref}
\bibliographystyle{icml2021}


\appendix

\onecolumn
\section*{\Large Appendices}

\section{Pool of Agents}\label{app:pool}
Here is the cross-play matrix of all the 100 agents trained with different MARL algorithms. There are five types of architectures (Table~\ref{tab:exact-architectures}) with two different seeds per MARL algorithm.

\begin{figure}[th!]
\centering
\includegraphics[width=.9\linewidth]{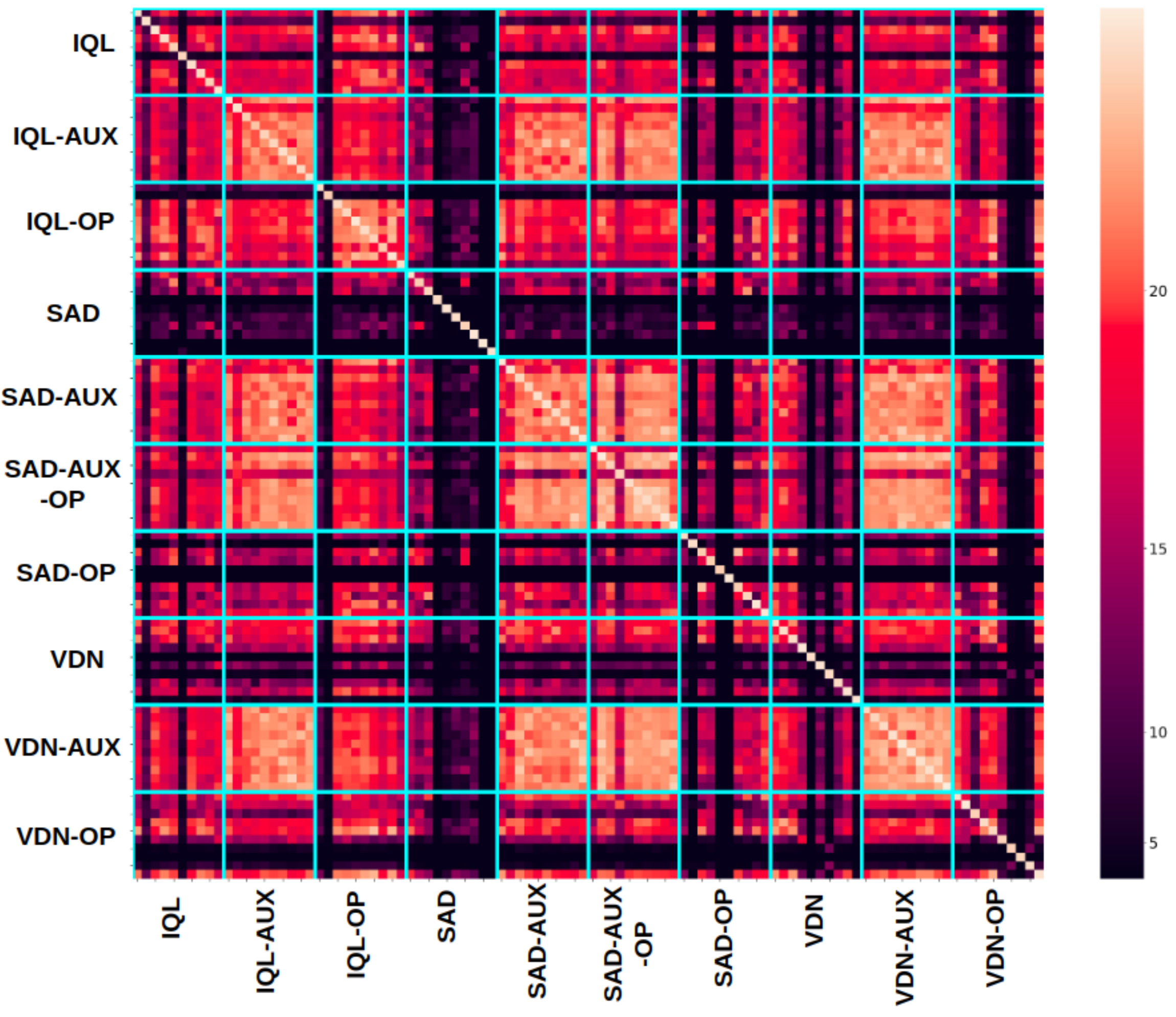}
\caption{The pool of 100 agents pre-trained through Self-Play using different MARL methods (IQL/VDN/OP/AUX/SAD, and their combinations). 10 agents having 5 different architectures with 2 seeds are generated with each of these MARL methods. $(i, j)_{th}$ element is the average score of agent $i$ paired with $j$ over 5k games. The diagonal entries indicate SP scores.}
\label{fig:matrix100} 
\end{figure}

\newpage

\section{List of agents}\label{app:agents_list}
In this section, we present the exact type of agents that we use as the \textit{learner} and its \textit{partners} in both \textit{easy} and \textit{hard} settings, as well as the set of 10 partners used in section~\ref{sec:LifelongRL} and in Appendix~\ref{app:all_LLL_benchmarks}. All these settings have an IQL agent of Type-2 as the \textit{learner} and a sequence of 5/10 agents (can be extended to any number of agents) as its partners. Table~\ref{tab:exact-architectures} has details of the exact architectures corresponding to these \textit{Types}. 

\begin{itemize}
    \item \textbf{Easy} : \textit{Learner} --- \Big\{ IQL (Type-2) \Big\} \\
 Partners --- \Big\{ IQL (Type-1), VDN (Type-3), VDN (Type-5), IQL+OP (Type-2), VDN+OP (Type-5) \Big\}.
 
 \item \textbf{Hard} : \textit{Learner} --- \Big\{ IQL (Type-2) \Big\} \\
Partners --- \Big\{ VDN+OP (Type-3), VDN (Type-4), VDN (Type-5),  IQL+OP (Type-3),  VDN (Type-3) \Big\}. \\
The partner agents in the Figure~\ref{fig:motivation} are these \textit{hard} agents. 

\item \textbf{10 agents}: \textit{Learner} --- \Big\{ IQL (Type-2) \Big\} \\
Partners --- \Big\{ VDN (Type-2), VDN (Type-3), IQL+OP (Type-2),  VDN+OP (Type-5),  IQL (Type-4), VDN+OP (Type-1), VDN (Type-4), IQL+OP (Type-3), VDN+OP (Type-1), VDN (Type-5) \Big\}. \\

\end{itemize}

The below are the set of 20 held-out agents that we use for across method evaluation in Tables~\ref{sota-marl-table_average} and~\ref{sota-marl-table-complete}.  \\

\textbf{Inter-CP} : \Big\{ IQL (Type-1), IQL (Type-3), IQL+OP (Type-4), IQL+OP (Type-5), VDN+AUX (Type-2), VDN+AUX (Type-3), SAD+OP (Type-3), SAD+OP (Type-1), SAD+OP+AUX (Type-3), SAD+OP+AUX (Type-1), SAD+AUX (Type-3), SAD+AUX (Type-1), SAD (Type-3), SAD (Type-2), IQL+AUX (Type-3), IQL+AUX (Type-1), VDN (Type-4), VDN (Type-2), VDN+OP (Type-5), VDN+OP(Type-4) \Big\}.

\begin{table*}[!ht]
\caption{Exact architectures used in the pool.}
\label{tab:exact-architectures}
\vskip 0.15in
\begin{center}
\begin{small}
\begin{sc}
\begin{tabular}{lcccccccccc}
\toprule
Agent  && RNN type & & Num of feed-forward layers &&  Num of RNN layers && RNN hid dim \\
\midrule
Type-1  && LSTM && 1 && 1 && 256 \\
Type-2  && LSTM && 2 && 2 && 256 \\
Type-3  && LSTM  && 1 && 2 && 512 \\
Type-4  && GRU &&  1 &&  2 &&  256 \\
Type-5  && GRU && 2 && 1 && 512 \\
\bottomrule
\end{tabular}
\end{sc}
\end{small}
\end{center}
\end{table*}

\section{LLL algorithms benchmarks}\label{app:all_LLL_benchmarks}

\begin{figure*}[h!]
\centering
\includegraphics[ width=.99\linewidth]{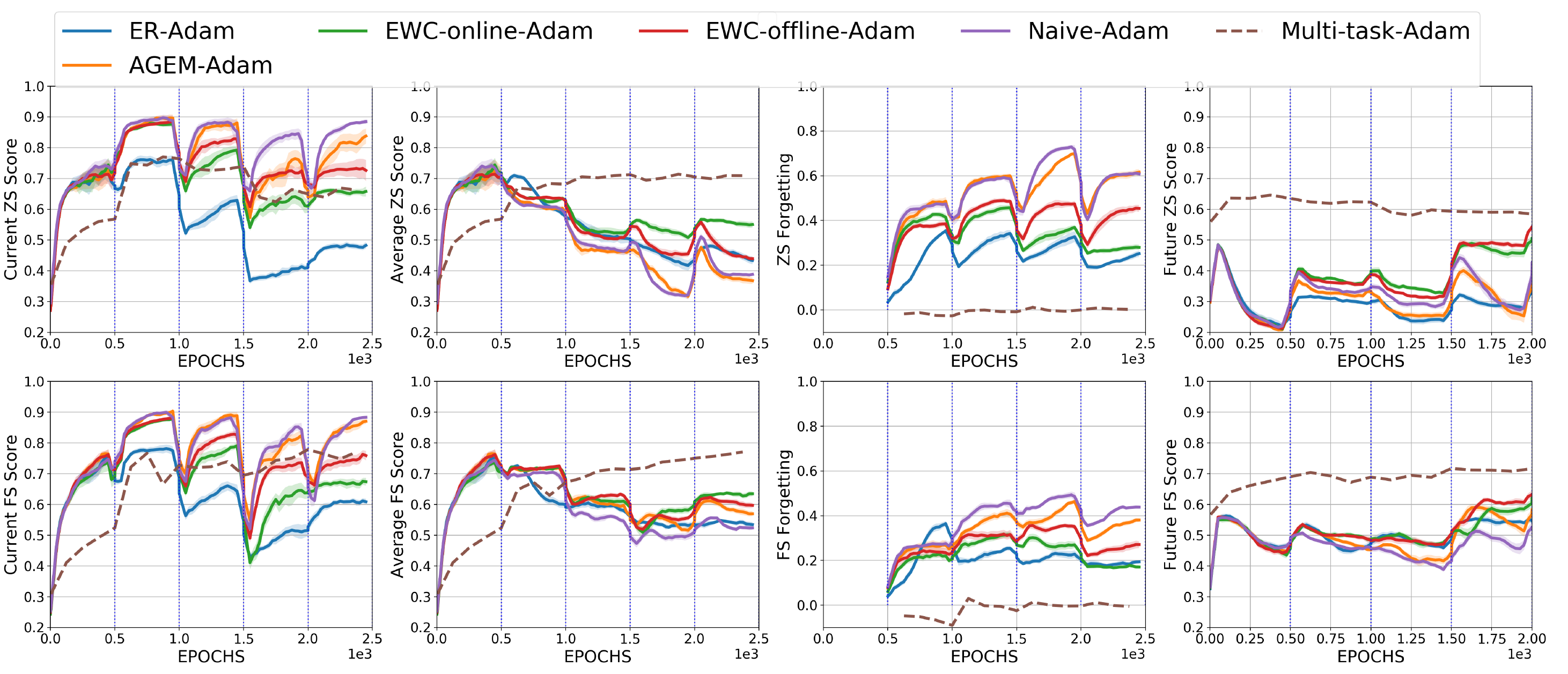}
\caption{Zero-shot (top row) and Few-shot (bottom row) performance of different LLL algorithms with Adam optimizer on \textit{hard} task. From left to right: current score, average score, forgetting and average future score respectively.}
\label{fig:LLLbenchmark_adam_hard_all} 
\end{figure*}

\begin{figure*}[h!]
\centering
\includegraphics[ width=.99\linewidth]{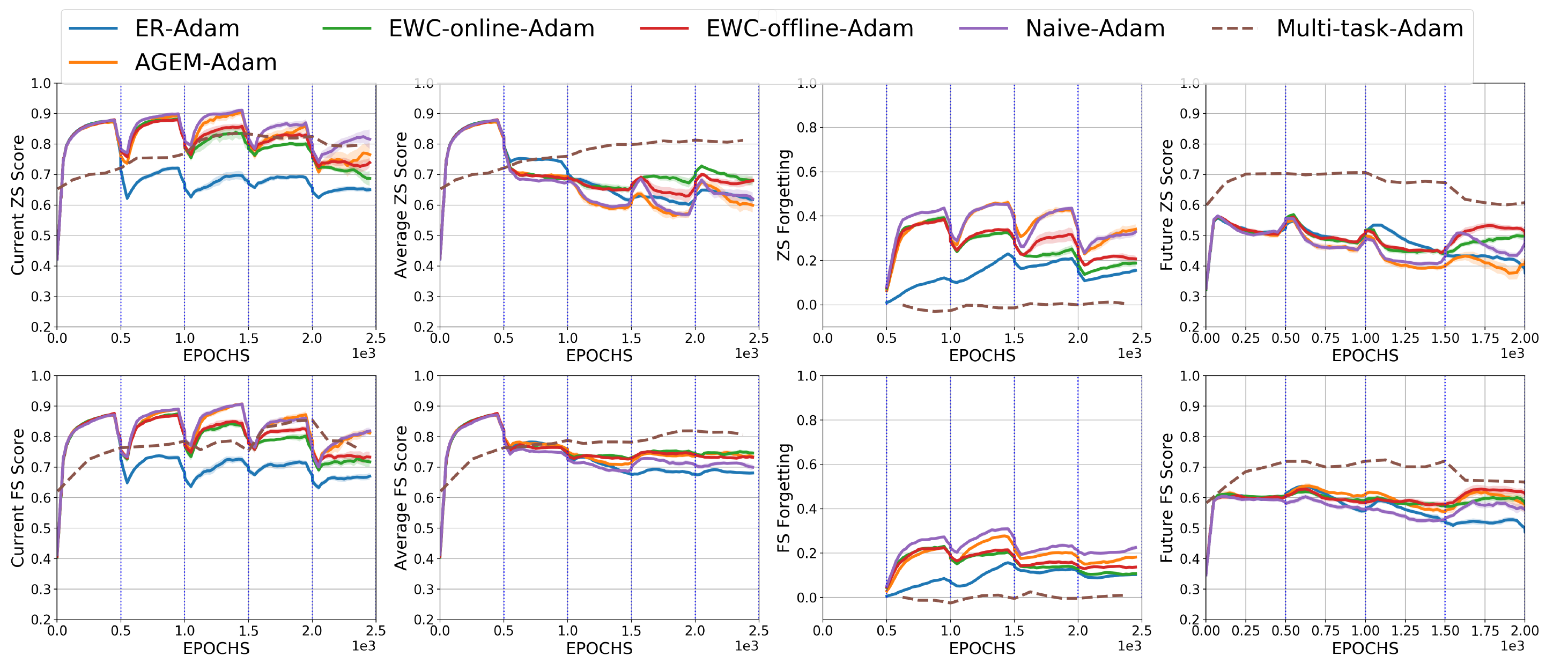}
\caption{Zero-shot (top row) and Few-shot (bottom row) performance of different LLL algorithms with Adam optimizer on \textit{easy} task. From left to right: current score, average score, forgetting and average future score respectively.}
\label{fig:LLLbenchmark_adam_easy_all} 
\end{figure*}

\begin{figure*}[h!]
\centering
\includegraphics[ width=.99\linewidth]{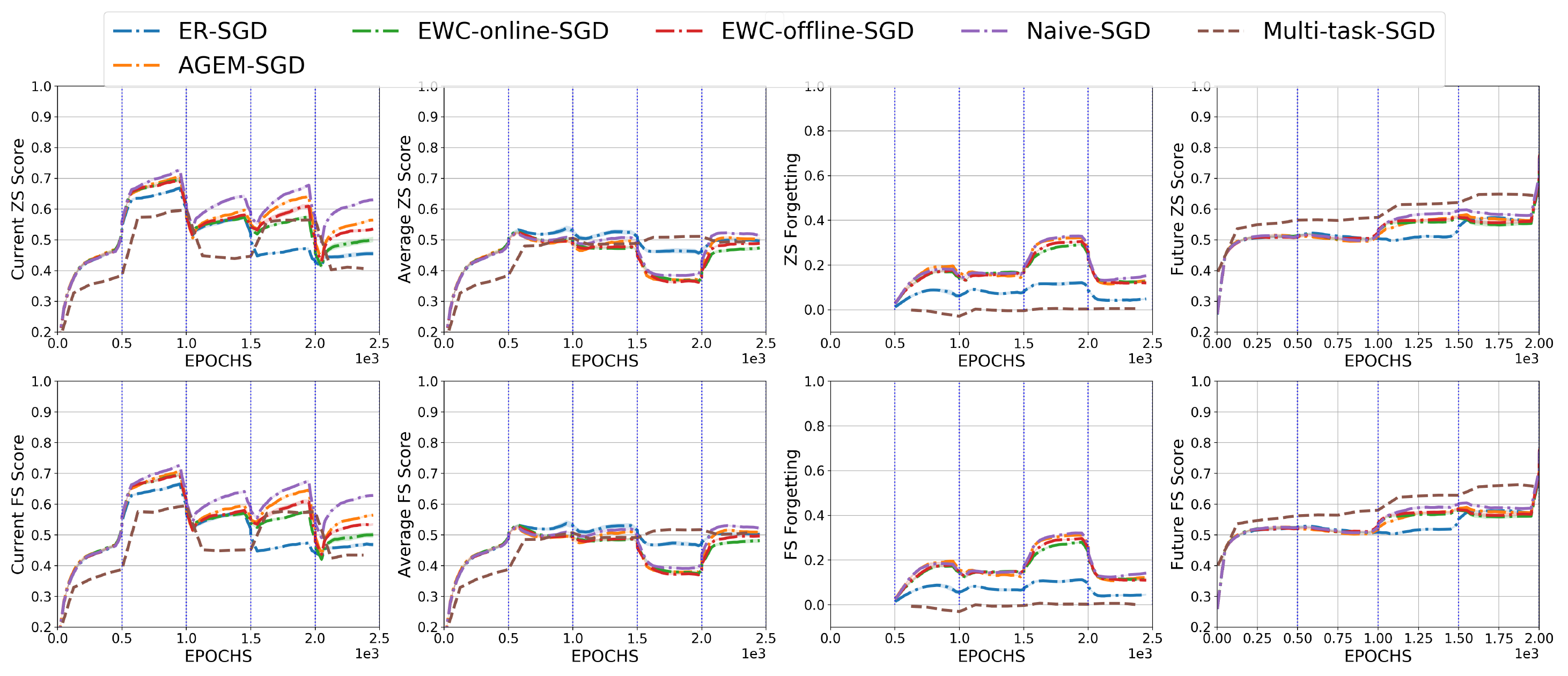}
\caption{Zero-shot (top row) and Few-shot (bottom row) performance of different LLL algorithms with SGD optimizer on \textit{hard} task. From left to right: current score, average score, forgetting and average future score respectively.}
\label{fig:LLLbenchmark_sgd_hard_all} 
\end{figure*}

\begin{figure*}[h!]
\centering
\includegraphics[ width=.99\linewidth]{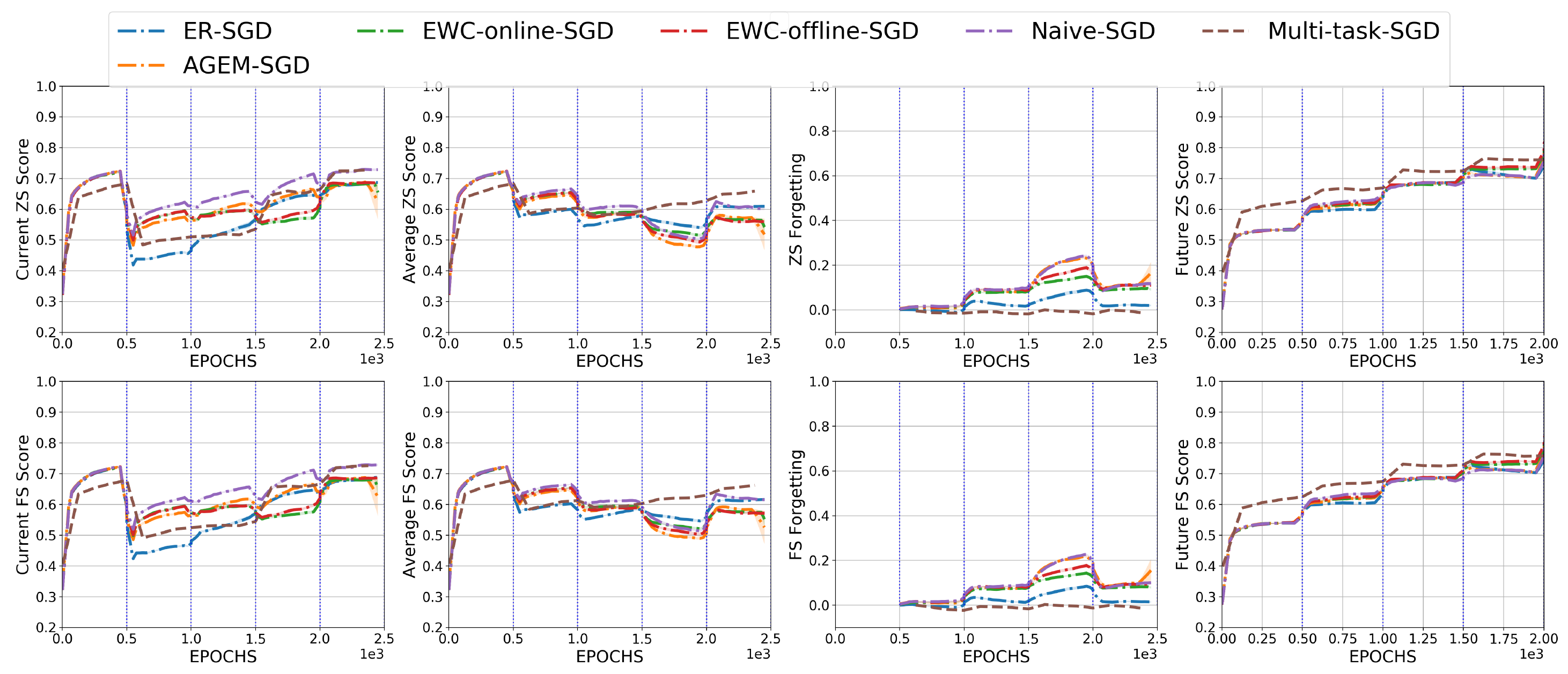}
\caption{Zero-shot (top row) and Few-shot (bottom row) performance of different LLL algorithms with SGD optimizer on \textit{easy} task. From left to right: current score, average score, forgetting and average future score respectively.}
\label{fig:LLLbenchmark_sgd_easy_all} 
\end{figure*}

\begin{figure*}[h!]
\centering
\includegraphics[trim=200 0 200 0, width=.99\linewidth]{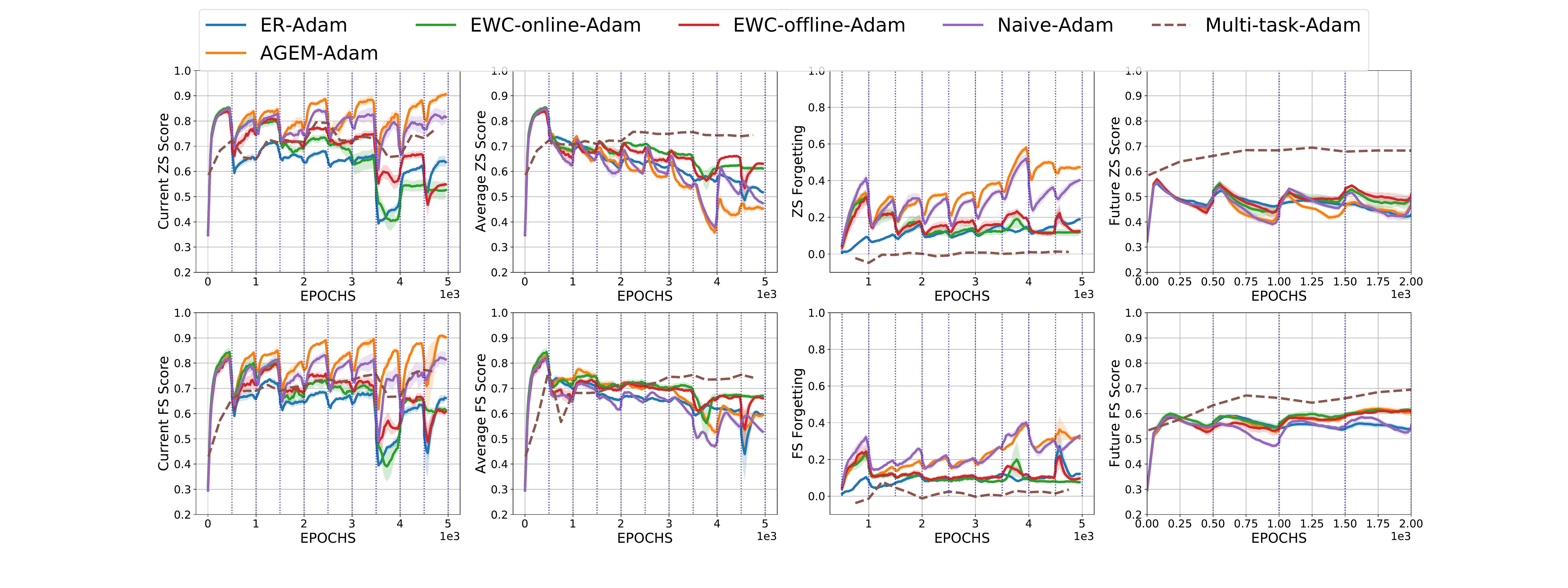}
\caption{Zero-shot (top row) and Few-shot (bottom row) performance of different LLL algorithms with Adam optimizer on 10 tasks. From left to right: current score, average score, forgetting and average future score respectively.}
\label{fig:LLLbenchmark_adam_10tasks_all} 
\end{figure*}

\begin{figure*}[h!]
\centering
\includegraphics[trim=200 0 200 0, width=.99\linewidth]{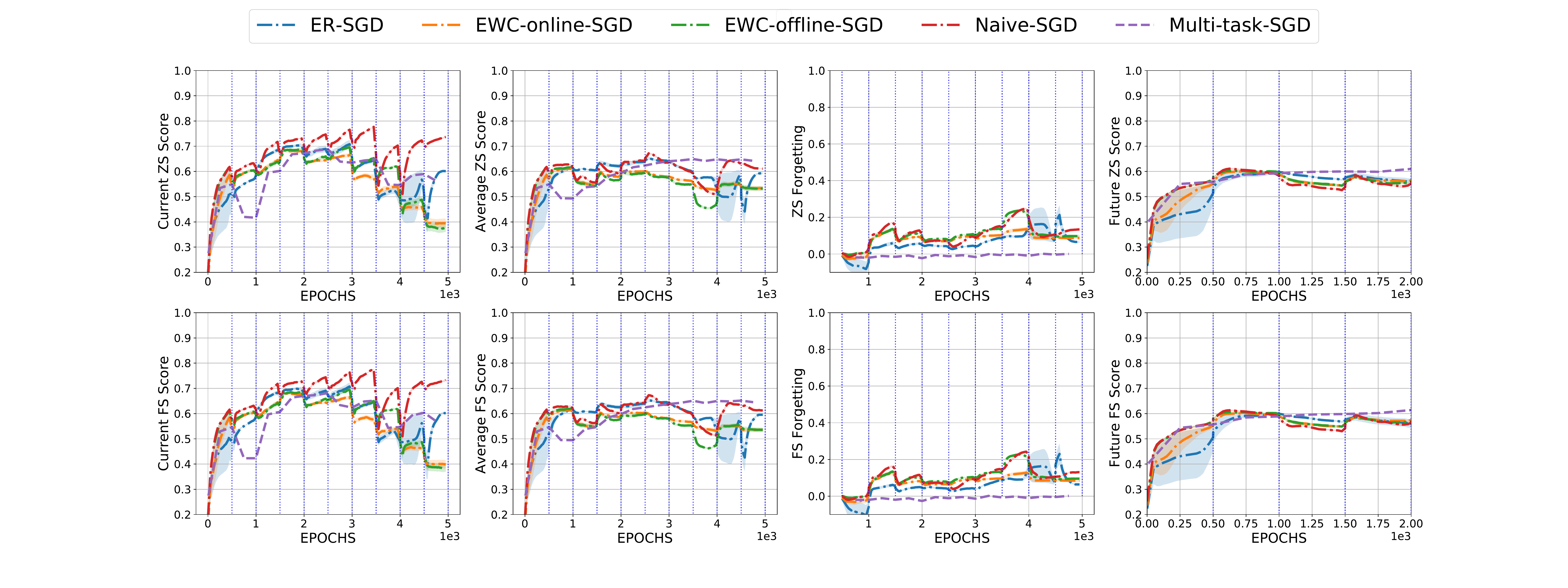}
\caption{Zero-shot (top row) and Few-shot (bottom row) performance of different LLL algorithms with SGD optimizer on 10 tasks. From left to right: current score, average score, forgetting and average future score respectively.}
\label{fig:LLLbenchmark_sgd_10tasks_all} 
\end{figure*}

\newpage

For Figures \ref{fig:LLLbenchmark_adam_hard_all}-\ref{fig:LLLbenchmark_sgd_easy_all}, the \textit{learner} is pre-trained with IQL method and is continually trained with either \textit{hard} or \textit{easy} agents mentioned in section~\ref{app:agents_list}. 

\newpage 

The sequential order of partners were chosen at random from the pre-trained pool in both \textit{easy} and \textit{hard} setting. Careful curation of the partner ordering and its effect on lifelong learning is left as future work.

\newpage
For Figures \ref{fig:LLLbenchmark_adam_10tasks_all}-\ref{fig:LLLbenchmark_sgd_10tasks_all}, the \textit{learner} is pre-trained with IQL method and is continually trained with 10 partners mentioned in section~\ref{app:agents_list}.
\newpage

\section{All hyperparameters and experiment details}

\begin{table*}[h]
\caption{All common hyperparameters and their description.}
\label{tab:hyperparameters}
\vskip 0.15in
\begin{center}
\begin{small}
\begin{sc}
\begin{tabular}{lcccccc}
\toprule
hyperparameters  && value & & Description \\
\midrule
$batchsize$ && 32 &&  batchsize used for both training \\
&& &&  and few-shot evaluation\\
\hline
$max \textunderscore train \textunderscore steps$ && 200m &&  maximum number of training steps per task\\
\hline
$max \textunderscore eval \textunderscore steps$ && 500k &&  maximum number of training steps\\
&& &&  during few-shot evaluation\\
\hline
$burn \textunderscore in \textunderscore frames$ && 10k &&  number of samples used to warm-up replay buffer\\
\hline
$ eval \textunderscore burn \textunderscore in \textunderscore frames$ && 1k &&  number of samples used to warm-up \\
&& &&  evaluation replay buffer\\
\hline
$replay \textunderscore buffer \textunderscore size$ && 32768 && replay buffer size during continual training \\
\hline
$ eval \textunderscore replay \textunderscore buffer \textunderscore size$ && 10000 &&  replay buffer size for few-shot evaluation\\
\hline
$ epoch \textunderscore len\textunderscore size$ && 200 && number of gradient updates per epoch \\
\hline
$ eval \textunderscore epoch \textunderscore len\textunderscore size$ && 50 && number of gradient updates for few-shot evaluation \\ 
\hline
$ eval \textunderscore freq$ && 25 && Learner is evaluated after each 25 epochs \\
\hline
$num \textunderscore thread$ && 10 &&  number of threads used for R2D2 actors\\
\hline
$num \textunderscore game \textunderscore per \textunderscore thread$ && 80 &&  number of game per threads used for R2D2 actors\\
\hline
$eval \textunderscore num \textunderscore thread$ && 10 && number of threads used for R2D2 actors\\
&& &&  during few-shot evaluation \\
\hline
$eval \textunderscore num \textunderscore game \textunderscore per \textunderscore thread$ && 10 && number of games per threads used for R2D2 actors \\
&& && during few-shot evaluation \\
\hline
$sgd \textunderscore momentum$ && 0.8 &&  momentum for SGD optimizer\\
\bottomrule
\end{tabular}
\end{sc}
\end{small}
\end{center}
\vskip -0.1in
\end{table*}

\begin{table*}[!ht]
\caption{Specific hyperparameters to each algorithm and their description}
\label{tab:specific-hyperparameters}
\vskip 0.15in
\begin{center}
\begin{small}
\begin{sc}
\begin{tabular}{lcccccc}
\toprule
hyperparameters  && value & & Description \\
\midrule
$ewc \textunderscore lambda$ && 50000 && EWC \\
$ewc \textunderscore gamma$ && 1 && EWC \\
$replay \textunderscore buffer \textunderscore size$ && 163840 && Multi-task \\
\bottomrule
\end{tabular}
\end{sc}
\end{small}
\end{center}
\vskip -0.1in
\end{table*}

\newpage

\section{MARL algorithms benchmarks}
In order to obtain the \textit{Intra-CP} scores for the existing MARL methods in the Table~\ref{sota-marl-table} and Table~\ref{sota-marl-table-complete} (referenced as BEST in caption), we take the agent from each training method that performs best with the \textbf{Inter-CP} agents listed above in section~\ref{app:agents_list} and evaluate them with the other 9 agents of the same method from the pretrained pool (Figure~\ref{fig:matrix100}).  However, in order to obtain the \textit{Intra-CP} scores for each MARL method in the Table~\ref{sota-marl-table_average} (referenced as AVG in caption), we pick one agent, evaluate it with the rest (barring itself) and repeat the same for all other agents. The average of these scores are reported. A similar process is followed for reporting \textit{Inter-CP} scores. The method of evaluating our LLL methods remains consistent in all the Tables(~\ref{sota-marl-table},~\ref{sota-marl-table_average},~\ref{sota-marl-table-complete}). For IQL+ER, we start with the IQL agent that has the least cross-play score and train it with \textit{Hard} agents sequentially using ER algorithm. In the case of IQL+AUX+ER, we start with an IQL agent that is pre-trained with AUX and is continually trained with the \textit{Hard} agents using ER algorithm. This continually trained agent is then evaluated with 9 other agents in either IQL or IQL+AUX respectively in order to obtain \textit{Intra-CP} scores. However, please note that the auxiliary task is used only during pre-training and is not used during continual training. Note that the middle row in the Table~\ref{sota-marl-table-complete} is generated using the latest models released by~\cite{hu2020other}.


\begin{table*}[!h]
 \caption{AVG : Comparison with other MARL algorithms on self-play (SP), cross-play evaluation scores within method (\textit{Intra-CP}), and across different methods (\textit{Inter-CP}). C: centralized training, GA: agents share their greedy action along with their standard action, L: true labels of cards needed, SYM: symmetries of the game needed upfront, P: require access to some pre-trained agents in sequence, UP: Having access to all the fixed pre-trained agents at the same time. ($\uparrow$ / $\downarrow$ = Difference in score after continual training \textcolor{red}{red}: pre-trained with MARL method, \textcolor{blue}{blue}: trained continually with LLL method)}
\label{sota-marl-table_average}
\vskip 0.15in
\begin{center}
\begin{small}
\begin{sc}
\begin{tabular}{lcccccccccc}
\toprule
Training method && SP & & Intra-CP &&  Inter-CP && Limitations \\
\midrule
\textcolor{red}{SAD}     && 23.78 $\pm$ 0.03 & & 4.38 $\pm$ 0.66 && 8.40 $\pm$ 0.23 && C+GA \\
\textcolor{red}{SAD+AUX}  && 23.82 $\pm$ 0.02 & & 21.15 $\pm$ 0.26 & & 17.01 $\pm$ 0.22 &&   C+GA+L\\
\textcolor{red}{SAD+OP}  && 23.67 $\pm$ 0.03 & & 12.00 $\pm$ 0.86 & & 12.79 $\pm$ 0.24 && C+Sym+GA \\
\textcolor{red}{SAD+AUX+OP}  && 23.88 $\pm$ 0.03 & & 22.01 $\pm$ 0.03 &&  17.08 $\pm$ 0.22 &&  C+Sym+L+GA       \\
\midrule
\textcolor{red}{IQL} + \textcolor{blue}{ER}  && 20.91 $\pm$ 0.05 ($\downarrow$ 2.98) && 15.73$\pm0.39$ (\textbf{$\uparrow$ 7.06})  && 16.32$\pm$0.21 (\textbf{$\uparrow$ 8.09})  && P \\
\textcolor{red}{IQL+AUX} + \textcolor{blue}{ER}  &&  22.34$\pm$ 0.06 ($\downarrow 1.46$)  && 20.90$\pm$ 0.06 ($\downarrow$0.15)  && \textbf{19.17$\pm$0.22}(\textbf{$\uparrow$1.33})  && L+P \\
\textcolor{red}{IQL} + \textcolor{blue}{Multi-task}   && 20.93$\pm0.09$ ($\downarrow 2.96$)  && 16.05$\pm$ 0.30(\textbf{$\uparrow$ 7.38})  && \textbf{17.88$\pm0.17$ ($\uparrow$ 9.65)}  &&  UP   \\
\bottomrule
\end{tabular}
\end{sc}
\end{small}
\end{center}
\vskip -0.1in
\end{table*}

\begin{table*}[!ht]
\caption{BEST : Comparison with other MARL algorithms on self-play (SP), cross-play evaluation scores within method (\textit{Intra-CP}), and across different methods (\textit{Inter-CP}). C: centralized training, GA: agents share their greedy action along with their standard action, L: true labels of cards needed, SYM: symmetries of the game needed upfront, P: require access to some pre-trained agents in sequence, UP: Having access to all the fixed pre-trained agents at the same time. ($\uparrow$ / $\downarrow$ = Difference in score after continual training, \textcolor{red}{red}: pre-trained with MARL method, \textcolor{blue}{blue}: trained continually with LLL method, $^*$ : results obtained using models released by~\citep{hu2020other})}
\label{sota-marl-table-complete}
\vskip 0.15in
\begin{center}
\begin{small}
\begin{tabular}{lcccccccc}
\toprule
Training Method && SP & & Intra-CP &&  Inter-CP & Limitations \\
\midrule
\textcolor{red}{SAD}     && $23.85 \pm 0.03$ & & $7.70 \pm 0.69$ && $14.60 \pm 0.24$ & C + GA \\
\textcolor{red}{SAD} + \textcolor{red}{AUX}  && $23.57 \pm 0.03$ & & $20.97 \pm 0.80$ & & $18.51 \pm 0.23$ &   C + GA + L\\ \textcolor{red}{SAD} + \textcolor{red}{OP}  && $24.14 \pm 0.03$ & & $10.10 \pm 0.87$ & & $16.09 \pm 0.25$ & C + Sym + GA \\
\textcolor{red}{SAD} + \textcolor{red}{AUX} + \textcolor{red}{OP}  && $23.40 \pm 0.04$ & & $21.23 \pm 0.25$ &&  17.77 $\pm$ 0.23 &  C + Sym + L + GA       \\
\midrule
$\textcolor{red}{SAD}^{*}$     && $23.97 \pm 0.04$ & & $2.52 \pm 0.0.34$ && $11.46 \pm 0.35$ & C + GA \\
$\textcolor{red}{SAD} + \textcolor{red}{AUX}^*$  && $24.09 \pm 0.03$ & & $17.65 \pm 0.69$ & & $17.60 \pm 0.42$ &   C + GA + L\\ 
$\textcolor{red}{SAD} + \textcolor{red}{OP}^*$  && 23.93 $\pm$ 0.02 & & 15.32 $\pm$ 0.65 & & 17.50 $\pm$ 0.34 & C + Sym + GA \\
$\textcolor{red}{SAD} + \textcolor{red}{AUX} + \textcolor{red}{OP}^*$  && $24.06 \pm 0.02$ & & $22.07 \pm 0.11$ &&  $17.45 \pm 0.38$ &  C + Sym + L + GA       \\
\midrule
\textcolor{red}{IQL} + \textcolor{blue}{ER}  && $20.91 \pm 0.05$ ($\downarrow 2.98$) && $15.73 \pm 0.39$ ($\mathbf{\uparrow 7.06}$)  && $16.32 \pm 0.21$ ($\mathbf{\uparrow 8.09}$)  & P \\
\textcolor{red}{IQL} + \textcolor{red}{AUX} + \textcolor{blue}{ER}  &&  $22.34 \pm 0.06$ ($\downarrow 1.46$)  && $20.90 \pm 0.06$ ($\downarrow 0.15$)  && $\mathbf{19.17 \pm 0.22}$ ($\mathbf{\uparrow 1.33}$)  & L + P \\
\textcolor{red}{IQL} + \textcolor{blue}{Multi-task}   && $20.93 \pm 0.09$ ($\downarrow 2.96$)  && $16.05 \pm 0.30$ ($\mathbf{\uparrow 7.38}$)  && $\mathbf{17.88 \pm 0.17}$ (\textbf{$\uparrow$ 9.65})  &  UP  \\
\bottomrule
\end{tabular}
\end{small}
\end{center}
\vskip -0.2in
\end{table*}

\end{document}